%% file: CastFlow.tex
\RequirePackage{fix-cm}
\documentclass[twocolumn]{svjour3}
\smartqed
\usepackage{amsmath,amsfonts}
\makeatletter
\@fleqnfalse
\makeatother
\usepackage{algorithm}
\usepackage{algpseudocode}
\usepackage{array}
\usepackage{textcomp}
\usepackage{xurl}
\providecommand{\doi}[1]{}
\renewcommand{\doi}[1]{\url{https://doi.org/#1}}
\usepackage{verbatim}
\usepackage{graphicx}
\usepackage{booktabs}
\usepackage{multirow}
\usepackage{microtype}
\usepackage{placeins}
\usepackage{flushend}
\usepackage{hyperref}
\usepackage{orcidlink}
\ExplSyntaxOn
\NewDocumentCommand{\biburl}{m}
  {
    \href{#1}
      {
        \int_compare:nNnTF { \str_count:n {#1} } > {16}
          {
            \exp_args:Nx \nolinkurl { \str_range:nnn {#1} {1} {-17} }
            \nobreak
            \mbox{\exp_args:Nx \nolinkurl { \str_range:nnn {#1} {-16} {-1} }}
          }
          { \mbox{\nolinkurl{#1}} }
      }
  }
\ExplSyntaxOff
\hypersetup{
    colorlinks=false,
    pdfborder={0 0 1},
    linkbordercolor={1 0 0},
    citebordercolor={0 1 0},
    urlbordercolor={0 1 1}
}
\journalname{}
\renewcommand{\makeheadbox}{}

\begin{document}
\flushbottom

\title{CastFlow: Bridging Evidence Acquisition and Accurate Prediction in Agentic Time Series Forecasting}

\author{Bokai~Pan\textsuperscript{\orcidlink{0009-0001-4496-8166}} \and
Mingyue~Cheng\textsuperscript{\orcidlink{0000-0001-9873-7681}} \and
Xiaoyu~Tao\textsuperscript{\orcidlink{0009-0000-0634-6254}} \and
Zhiding~Liu\textsuperscript{\orcidlink{0000-0003-0994-473X}} \and
Shuo~Yu\textsuperscript{\orcidlink{0009-0006-1060-5451}} \and
Yuchong~Wu\textsuperscript{\orcidlink{0009-0001-4389-9613}} \and
Qi~Liu\textsuperscript{\orcidlink{0000-0001-6956-5550}} \and
Defu~Lian\textsuperscript{\orcidlink{0000-0002-3507-9607}} \and
Enhong~Chen\textsuperscript{\orcidlink{0000-0002-4835-4102}}}
\authorrunning{Pan et al.}
\titlerunning{CastFlow: Bridging Evidence Acquisition and Accurate Prediction}
\institute{Bokai~Pan, Mingyue~Cheng, Xiaoyu~Tao, Zhiding~Liu, Shuo~Yu, Yuchong~Wu, Qi~Liu, Defu~Lian, and Enhong~Chen \at
State Key Laboratory of Cognitive Intelligence, University of Science and Technology of China, Hefei 230026, China.\\
\email{mycheng@ustc.edu.cn} (Mingyue~Cheng, corresponding author for this manuscript).}
\date{Preprint}

\maketitle

\begin{abstract}
Time series forecasting is a fundamental task in temporal data analytics, and large language model (LLM)-based agents offer a promising paradigm through autonomous tool interaction and evidence-grounded reasoning.
However, two fundamental challenges remain: how to acquire forecast-relevant evidence through adaptive tool interactions, and how to translate such evidence into accurate numerical predictions.
To address these challenges, we propose CastFlow, a role-specialized agentic forecasting framework that bridges general-purpose tool reasoning and specialized temporal forecasting.
Specifically, CastFlow introduces an Experience Memory that distills successful tool-use trajectories from training data, guiding a frozen LLM to orchestrate a multi-view toolkit for evidence acquisition.
A lightweight specialized LLM then performs evidence-grounded temporal reasoning to refine an ensemble forecasting prior.
To enhance this capability, we develop a two-stage training strategy combining supervised fine-tuning and reinforcement learning with verifiable rewards, where final forecasting outcomes of the overall workflow guide the optimization of the specialized model.
Extensive experiments on ten real-world benchmarks demonstrate that CastFlow outperforms competitive baselines on nine datasets.
Our findings highlight the importance of bridging autonomous evidence acquisition and specialized temporal reasoning for accurate and adaptive agentic forecasting.
The source code is available at \url{https://github.com/Forever-Pan/CastFlow}.
\end{abstract}

\section{Introduction}
Time series forecasting (TSF) supports data-driven decisions in energy and water systems, with applications including renewable energy generation forecasting~\cite{xfyun_renewable_power_challenge_2025} and streamflow prediction~\cite{schaake2006us}.
Accurate forecasting is challenging because temporal dynamics evolve, dependencies vary, and historical patterns can change abruptly~\cite{li2018diffusion,zhao2024lift,benidis2022deep}.
Conventional forecasting models primarily learn mappings from past observations to future values~\cite{cheng2025comprehensive,qiu2024tfb}.
The emergence of large language model (LLM) agents creates an opportunity to augment these mappings with active investigation: agents can inspect temporal patterns, invoke analytical tools, and incorporate diagnostic evidence into forecasting decisions~\cite{yao2023react,cheng2026position}.
Viewed as an analytical workflow, agentic forecasting must coordinate historical data, tool execution, and reusable experience, connecting predictive modeling with data access and model management~\cite{fischer2013integrated,januschowski2024stack}.

\begin{figure*}[!t]
    \centering
    \includegraphics[width=0.90\linewidth]{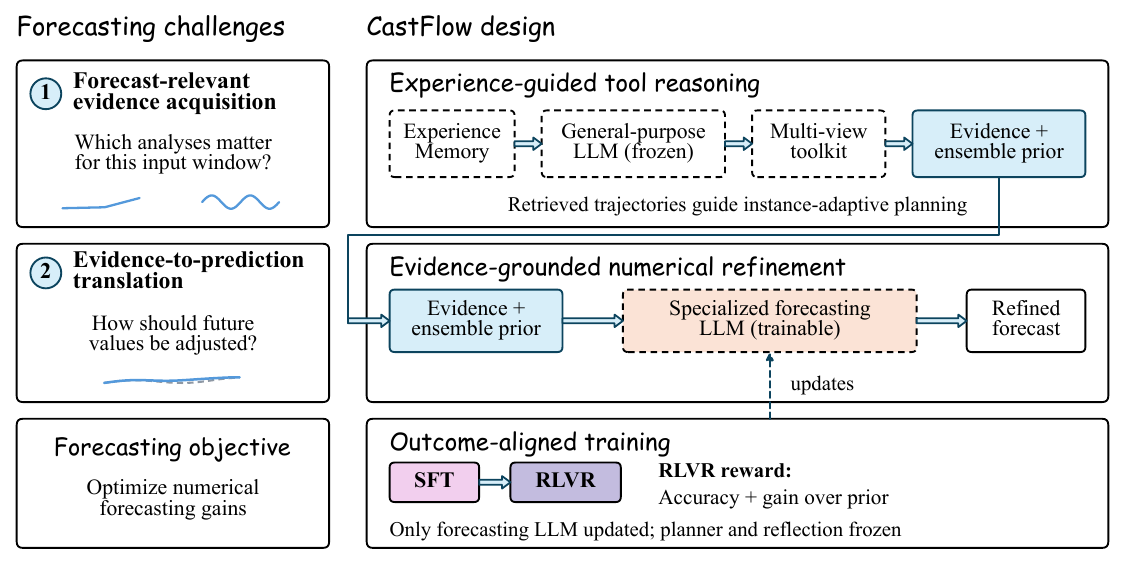}
    \caption{Conceptual motivation for CastFlow. Experience-guided tool reasoning addresses forecast-relevant evidence acquisition, while a specialized LLM translates diagnostic evidence and an ensemble prior into refined numerical forecasts. SFT followed by RLVR updates only the forecasting LLM; the RLVR reward combines absolute prediction accuracy with improvement over the prior, while planning and reflection \mbox{remain frozen}}
    \label{fig:example}
\end{figure*}

Recent studies have extended LLM-based forecasting through inference-time temporal reasoning, reinforcement fine-tuning, and experience-conditioned reasoning~\cite{timereasoner,time_r1,tao2026memcast}.
Tool-assisted frameworks further support diagnostic analysis, reflective revision, and learned sequential tool-use policies~\cite{tsscientist,alphacast,tao2026cast}.
These advances enable forecasting workflows to investigate and revise predictions, but tool access and multi-step reasoning alone do not determine which evidence is useful or how it should alter future numerical values.
Our focus is the coordination between experience-guided evidence acquisition and a selectively trained numerical forecaster: retrieved procedures guide what to inspect, while forecasting supervision teaches how to use the resulting context.
This motivates our central question: \emph{How can LLM agents acquire forecast-relevant evidence and translate it into accurate time series predictions?}

Two linked challenges make this translation difficult.
The first is \emph{forecast-relevant evidence acquisition}: an agent must identify which observations and analyses matter for the temporal dynamics of the current instance.
For example, an electricity-price sequence undergoing a recent trend shift may call for different diagnostics from one dominated by recurring seasonal patterns.
Applying the same tool plan to every instance can introduce irrelevant evidence and unnecessary computation, whereas planning anew without task experience may overlook useful analyses.
The agent therefore needs guidance for adapting its investigation to the observed context, using only information available at the forecast origin.
The second challenge is \emph{evidence-to-prediction translation}: heterogeneous diagnostic signals must inform precise numerical values across the forecasting horizon.
Recognizing an upward trend or a possible turning point does not specify how much each future value should change.
General-purpose reasoning can organize such information, but learning how it should inform numerical adjustments calls for adaptation to \mbox{forecasting-specific objectives}.

Our key insight is that evidence acquisition and numerical refinement can use different forms of \mbox{adaptation}.
Retrieved procedural experience guides a general-purpose LLM in adapting tool plans to the current instance, while forecasting-specific supervision adapts a specialized model's parameters to numerical prediction.
We therefore propose \emph{CastFlow}, a role-specialized agentic framework that bridges evidence acquisition and accurate prediction through complementary model responsibilities.
A frozen general-purpose LLM plans tool use and checks forecasts for consistency with the acquired evidence, providing feedback for bounded revision.
A lightweight specialized LLM uses the diagnostic context and ensemble forecasting prior to generate refined numerical forecasts.
Figure~\ref{fig:example} illustrates how these complementary roles address the two challenges, supported by \mbox{outcome-aligned training}.

To address evidence acquisition, CastFlow builds an \emph{Experience Memory} from training-time tool-use trajectories.
For each training instance, candidate exploration paths are evaluated by horizon-wide forecasting error; the path with the lowest MSE among those passing format and logic checks is retained as structured experience containing the tool schedule, diagnostic observations, and final forecasting response.
The memory is fixed before test inference, and the observed input window retrieves relevant entries to guide the frozen planner.
Retrieved procedures inform the choice of optional diagnostics; the ensemble prior and exogenous analysis are always included.
The multi-view toolkit supplies an ensemble prior and complementary evidence on statistical and spectral structure, local temporal dynamics, and cross-variable dependencies.
Tool execution grounds this evidence in information available at the forecast origin, including previously issued auxiliary \mbox{forecasts where applicable}.

To address evidence-to-prediction translation, the specialized LLM conditions its forecasting response on the acquired diagnostics and ensemble prior.
The prior supplies a quantitative starting point, and diagnostic evidence informs adjustments to its temporal trajectory; the model generates the complete refined forecast conditioned on this context.
We develop a two-stage training strategy combining supervised fine-tuning (SFT) with reinforcement learning using verifiable rewards (RLVR).
SFT teaches the forecasting response protocol and the use of diagnostic context in temporal reasoning and numerical generation.
RLVR then optimizes the specialized model using verifiable rewards computed from its final forecasting response under workflow-generated context, combining absolute prediction accuracy with improvement over the ensemble prior.
Only the specialized forecasting model is optimized; the general-purpose planning and reflection model remains frozen \mbox{throughout both stages}.

We evaluate CastFlow on ten real-world benchmarks covering multiple application domains and short- and long-term forecasting settings.
Among the evaluated methods, CastFlow achieves the lowest mean squared error (MSE) and mean absolute error (MAE) on nine datasets.
Component and training ablations examine forecasting performance, while separate execution studies evaluate diagnostic feature reuse and controlled interruption recovery.
These findings support combining experience-guided evidence acquisition with specialized, outcome-aligned numerical refinement to improve \mbox{forecasting accuracy}.

Our main contributions are as follows:
\begin{itemize}
    \item We propose CastFlow, a role-specialized agentic framework that couples experience-guided evidence acquisition with prior-conditioned numerical refinement, using a frozen planning and reflection model and a selectively trained forecasting LLM.
    \item We construct a retrievable memory by selecting valid training trajectories according to forecasting error, and use it to guide instance-adaptive tool planning. SFT initializes the forecasting model, and RLVR rewards absolute accuracy and improvement over the ensemble prior.
    \item CastFlow achieves the lowest MSE and MAE on nine of ten real-world benchmarks among the evaluated methods. Component and training ablations, memory analyses, and controlled execution studies characterize its forecasting and workflow behavior.
\end{itemize}

\section{Related Work}
\subsection{Traditional Time Series Forecasting}
Time series forecasting spans statistical, machine learning, deep learning, and foundation models~\cite{qiu2024tfb,cheng2025comprehensive,fang2026pipeline}.
Classical approaches such as ARIMA~\cite{ARIMA} and ETS~\cite{gardner1985exponential} characterize trend, seasonality, and temporal dependence through explicit structural assumptions, offering interpretable and efficient forecasting baselines~\cite{wang2024deep}.
Machine learning methods extend these baselines using lagged observations and engineered features, including support vector regression~\cite{sapankevych2009time}, boosting~\cite{chen2016xgboost,ke2017lightgbm}, and feature-based strategies~\cite{bontempi2013machine}.
Deep architectures reduce manual feature engineering and capture temporal patterns through diverse inductive biases~\cite{li2025hyperimts,huang2025timebase}, including linear, convolutional~\cite{eldele2024tslanet}, Transformer~\cite{chen2025closer,zhou2022fedformer}, and state-space models~\cite{ahamed2024timemachine}, with representative methods such as N-HiTS~\cite{challu2023nhits}, ETSformer~\cite{woo2022etsformer}, iTransformer~\cite{liu2023itransformer}, and ConvTimeNet~\cite{cheng2025convtimenet}.
Chronos~\cite{ansari2024chronos}, TimesFM~\cite{das2023timesfm}, and Sundial~\cite{liu2025sundial} extend forecasting through large-scale pretraining with promising zero-shot and cross-domain capabilities, while ZooCast~\cite{shi2025zoocast} explores model-zoo selection for zero-shot prediction.
These methods typically emphasize predictive modeling from historical observations, with limited support for test-time tool interaction, explicit evidence acquisition, and iterative forecast revision~\cite{cheng2026position}.

\subsection{LLM-Based Time Series Forecasting}
Research on LLM-based time series modeling spans multiple tasks and adaptation strategies~\cite{zhang2024large,jiang2024empowering}.
In forecasting, prominent approaches include prompt reformulation, input reprogramming, and semantic alignment~\cite{liu2025timecma,qiu2027east}.
PromptCast~\cite{xue2023promptcast} reformulates numerical sequences as textual prompts, Time-LLM~\cite{jin2023time} reprograms frozen LLMs with patch-level temporal embeddings, and S$^2$IP-LLM~\cite{pan2024s2ip} and TokenCast~\cite{tao2025tokencast} explore semantic alignment and token-based modeling.
Numerical tokenization and uncertainty calibration also affect pretrained LLM forecasts~\cite{gruver2023large}.
Alongside these adaptations, TimeReasoner~\cite{timereasoner} studies slow-thinking temporal reasoning, while TimeSeriesScientist~\cite{tsscientist} develops agentic time series analysis. AlphaCast~\cite{alphacast} combines diagnostic context, clustered case-based model references, and reflective LLM forecast refinement.
Time-R1~\cite{time_r1} uses reinforcement fine-tuning to strengthen multi-step temporal reasoning, while Cast-R1~\cite{tao2026cast} learns tool-augmented sequential policies through SFT and multi-turn RL.
MemCast~\cite{tao2026memcast} organizes historical patterns, reasoning experience, and general laws into hierarchical memory to condition forecasting; FLAIRR-TS~\cite{jalori2025flairr} coordinates retrieval, forecasting, and prompt-refinement agents without weight updates.
These studies emphasize reasoning, interaction, and experience reuse~\cite{cheng2026position}.
CastFlow shares prior-guided numerical refinement with AlphaCast, but separates adaptation mechanisms: procedural memory guides tool planning, planning and reflection remain frozen, and SFT and RLVR update only the specialized forecasting model. This differs from Cast-R1's learned sequential tool-use policy and FLAIRR-TS's test-time \mbox{prompt optimization}.

\subsection{Tool Use, Memory, and Post-Training}
Toolformer~\cite{schick2023toolformer} shows how language models can learn to invoke external tools, ReAct~\cite{yao2023react} interleaves reasoning with actions, and DEPS~\cite{wang2024describeexplainplanselect} supports interactive planning in complex environments.
DocETL~\cite{shankar2025docetl} optimizes document pipelines through agent-guided rewriting and evaluation; LLM-PQA~\cite{li2024llmpqa} connects natural-language prediction queries with data lakes and model zoos for model retrieval or on-demand training.
For time series anomaly detection, AnomaMind~\cite{tao2026anomamind} combines tool-augmented reasoning with a structured workflow.
Self-Refine~\cite{madaan2023selfrefine} studies iterative refinement through self-feedback, while Reflexion~\cite{shinn2023reflexion} uses verbal reinforcement and episodic memory for self-correction across reasoning trajectories.
MemoTime~\cite{tan2026memotime} uses experience memory for temporal knowledge graph question answering, illustrating memory use beyond numerical forecasting.
STaR~\cite{zelikman2022star} studies self-improvement through generated reasoning traces, while DeepSeek-R1~\cite{guo2025deepseek} and Qwen3~\cite{yang2025qwen3technicalreport} develop reasoning capabilities through post-training that includes reinforcement learning.
In adjacent temporal tasks, TimeOmni-1~\cite{guan2025timeomni} studies reward-driven time series reasoning, whereas AlphaAgentEvo~\cite{tangalphaagentevo} applies self-evolving agentic RL to financial alpha mining.
CastFlow combines these mechanisms through complementary roles: a frozen LLM acquires diagnostic evidence, while a trainable forecasting model refines an ensemble~\mbox{forecast prior}.

\section{Preliminaries}
\label{sec:preliminaries}

\subsection{Problem Definition}
\label{sec:forecasting_task}
At forecast origin $t$, let $\mathbf x\in\mathbb R^{L\times C_x}$ denote the $L$-step history ending at $t$, comprising target observations and available covariates, and $\mathbf y\in\mathbb R^{H\times C_y}$ the next $H$ target steps. Optional covariate forecasts $\mathbf u^{[t]}\in\mathbb R^{H\times C_u}$ cover the same horizon and were issued by $t$. Here, $C_x$, $C_u$, and $C_y$ are the corresponding channel counts, with $C_y=1$ in the evaluated tasks. A numerical output of the complete forecasting procedure $F$ is
\begin{equation}
\hat{\mathbf y}=F(\mathbf x,\mathbf u^{[t]})
\in\mathbb R^{H\times C_y},
\label{eq:forecasting_task}
\end{equation}
with $\mathbf u^{[t]}$ omitted when unavailable. Forecast error is measured by MSE and MAE; model-specific training objectives are given in Section~\ref{sec:optimization}.

\subsection{Information Availability}
\label{sec:information_availability}
Let $a(v)$ denote the availability time of source value $v$. Admissible source information is
\begin{equation}
\mathcal I_t=\{v:a(v)\leq t\}.
\label{eq:available_information}
\end{equation}
Inputs must belong to $\mathcal I_t$; diagnostics are derived from these inputs and fixed offline resources. Supplied forecasts remain fixed within each request; later revisions and realized future covariates are excluded. At test time, model parameters and Experience Memory remain fixed, while request-local diagnostics and feedback may evolve. The current request's target sequence is used only for subsequent evaluation. Appendix~\ref{app:reproducibility} details offline target use and \mbox{dataset provenance}.

\section{Methodology}
\label{sec:method}
CastFlow connects experience-guided evidence acquisition with evidence-grounded numerical forecasting. We first describe its role-specialized architecture, then explain how historical experience guides tool planning, how a specialized model refines an ensemble forecasting prior, and how forecasting outcomes \mbox{supervise this model}.

\begin{figure*}[!t]
    \centering
    \includegraphics[width=\linewidth]{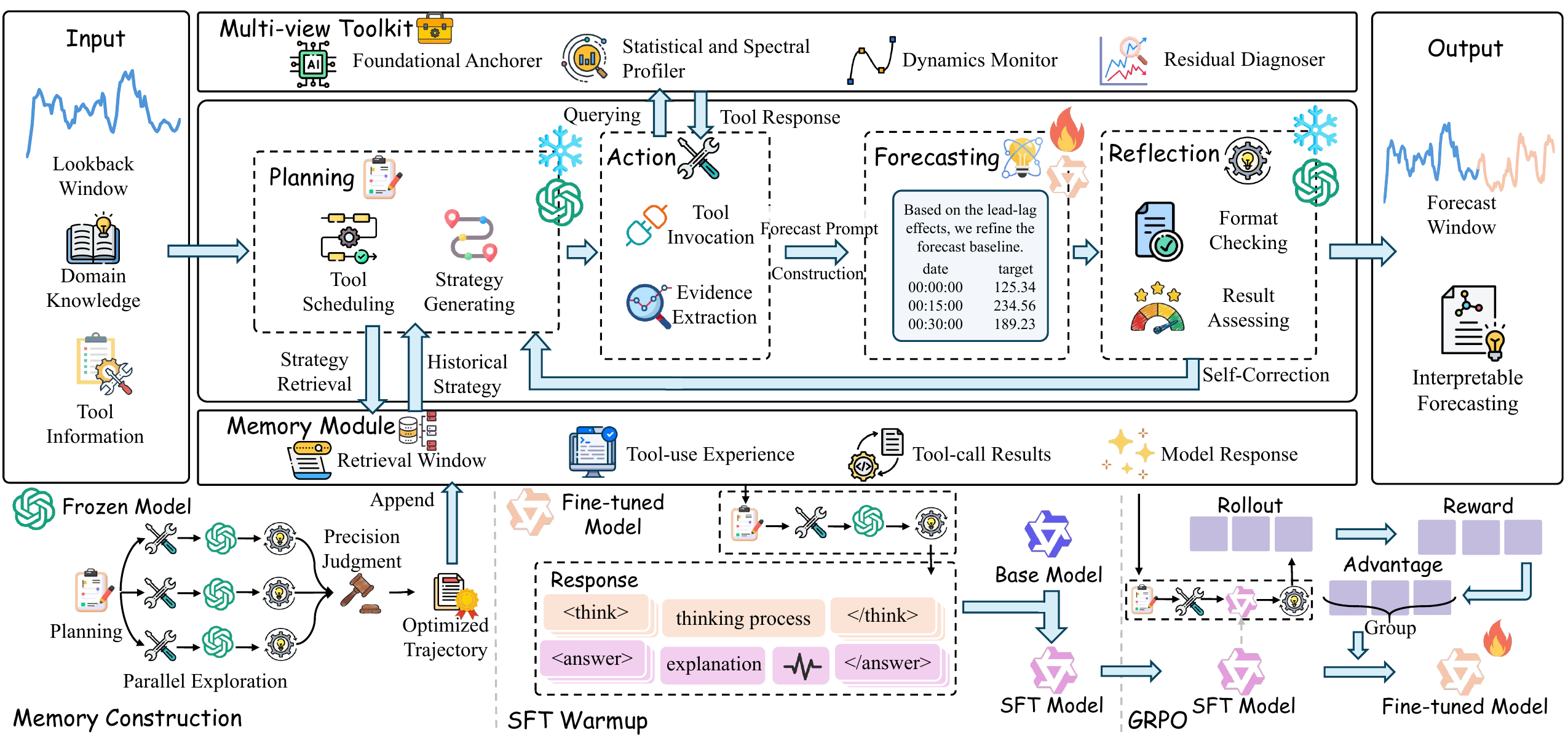}
    \caption{Overview of CastFlow. Planning and reflection use a frozen general-purpose LLM, while the specialized forecasting LLM refines the ensemble prior using acquired evidence. The Memory Module denotes training-built Experience Memory and is read-only at test time; the residual diagnoser executes only during training. Online inputs include origin-issued horizon covariate forecasts where supplied. Reflection checks format and evidence consistency. Offline SFT and response-level GRPO update only \mbox{the forecaster}}
    \label{fig:framework}
\end{figure*}

\subsection{Framework Overview}
\label{sec:framework_overview}
As illustrated in Fig.~\ref{fig:framework}, CastFlow comprises planning, action, forecasting, and reflection stages. A general-purpose LLM with frozen parameters $\Theta$ performs tool planning and evidence-consistency assessment. A lightweight forecasting LLM with trainable parameters $\theta$ generates numerical predictions from the resulting context. The action stage is a deterministic execution interface: it invokes the planned tools and returns their outputs without introducing another \mbox{trainable reasoning model}.

The framework separates offline preparation from online inference. Offline, candidate tool-use trajectories on training instances are evaluated against observed targets to construct an Experience Memory. Teacher-generated forecasting responses then support supervised fine-tuning (SFT), followed by reinforcement learning with verifiable rewards (RLVR). Online, the framework retrieves historical experience, plans and executes tools using information available at the forecast origin, and supplies the resulting diagnostic evidence and ensemble prior to the specialized forecaster. Reflection checks the response and can initiate bounded revision. Ground-truth forecasting targets are used for offline selection and training rewards, not for online \mbox{planning or verification}.

Let $\mathcal M$ denote the offline Experience Memory and $\mathcal M_{\mathrm{ret}}$ its retrieved entries. For an inference round $j$, the information flow is summarized by
\begin{equation}
\begin{aligned}
\mathcal M_{\mathrm{ret}} &= \operatorname{Retrieve}(\mathbf x,\mathcal M),\\
A_j &= \operatorname{Plan}_{\Theta}(\mathbf x,\mathcal M_{\mathrm{ret}},\mathcal H_j),\\
(\hat{\mathbf y}_{\mathrm{base},j},\mathcal D_j)
  &= \operatorname{Execute}(A_j,\mathbf x,\mathbf u^{[t]}),\\
r_j &\sim \pi_{\theta}(\,\cdot\mid\mathcal C_j).
\end{aligned}
\label{eq:castflow_dataflow}
\end{equation}
Here, $A_j$ is a tool schedule, $\mathcal H_j$ contains available workflow observations and feedback, and $\mathcal C_j$ is the forecasting context assembled from the lookback, origin-issued horizon covariate forecasts, ensemble prior, and diagnostic evidence. The response $r_j$ contains the forecasting model's reasoning and numerical output. The initial history is empty; a failed verification supplies feedback for a subsequent round. The roles of $\Theta$ and $\theta$ remain unchanged across these rounds. Sections~\ref{sec:evidence_acquisition}--\ref{sec:forecast_refinement} describe this inference process, and Section~\ref{sec:optimization} explains \mbox{selective training}.

\subsection{Experience-Guided Evidence Acquisition}
\label{sec:evidence_acquisition}
Evidence acquisition determines which observations accompany the numerical forecasting prior. CastFlow first retrieves structured experiences from training instances, uses them to guide tool planning, and executes the resulting schedule to obtain complementary diagnostic views. For a workflow round $j$, this stage produces the ensemble prior $\hat{\mathbf{y}}_{\mathrm{base}}$ and diagnostic evidence $\mathcal D_j$ for the \mbox{specialized forecaster}.

\label{sec:experience_memory}
For each training instance, CastFlow expands an initial plan into $K_{\mathrm{exp}}$ parallel exploration paths. Each path specifies a tool schedule, collects its outputs, and generates a candidate forecast. When at least one candidate passes both format and logic validation, we retain the valid path with the lowest horizon-wide MSE:
\begin{equation}
\tau^* = \underset{\tau\in\mathcal{T}_{\mathrm{valid}}}{\arg\min}
\frac{1}{H C_y}\sum_{h=1}^{H}
\left\|\hat{\mathbf{y}}_{t+h}^{(\tau)}-\mathbf{y}_{t+h}\right\|_2^2,
\label{eq:memory_selection}
\end{equation}
where $H$ is the forecasting horizon, $C_y$ is the number of target channels, and $\mathcal{T}_{\mathrm{valid}}$ contains the candidates accepted by both checks. Forecasting targets are used here to compare training paths. An archived entry is a structured tuple $e=\langle\mathbf{x},A^*,O^*,r^*\rangle$, containing the lookback window, selected tool schedule, corresponding tool outputs, and the final forecasting response $r^*$ of the selected trajectory $\tau^*$. Distillation thus means selecting and retaining structured experiences from the explored paths. Here, a successful trajectory means the valid, lowest-error candidate among the explored paths; selection does not impose an additional requirement that it outperform \mbox{the ensemble prior}.

The Experience Memory used in the main evaluation is constructed from training instances and their training targets, and is fixed before test inference. Test requests only retrieve existing records: their responses and targets are not written back to the memory, and test errors do not select or re-rank stored experiences. Append and merge concern the organization of training experience before this fixed-memory evaluation, rather than label-driven adaptation during testing. This separation also distinguishes persistent historical experience from the request-local observations and reflection feedback accumulated in $\mathcal H_j$.

Entries are indexed by vector similarity. For a new lookback window $\mathbf{x}$, retrieval returns up to $K$ highest-similarity entries satisfying $\operatorname{sim}(\mathbf{x},\mathbf{x}_e)\geq\eta$, where $\eta$ is the similarity threshold. These entries provide examples of tool use and its associated evidence; their training outcomes support historical selection without requiring the current forecasting target. Memory influences prediction through two routes: retrieved schedules guide the planner's choice of diagnostics, and retrieved procedural guidance also conditions the forecasting response. Stored observations and responses describe historical training instances; evidence about the current window is produced by its own tool executions. Experience Memory provides procedural context for planning and forecasting, whereas the anchorer's historical library supplies model-selection and weighting information for the ensemble prior in Section~\ref{sec:forecast_refinement}.

\paragraph{Experience-Guided Tool Planning}
\label{sec:tool_planning}
The frozen planner maps the input window and retrieved experiences $\mathcal M_{\mathrm{ret}}$ to a structured tool schedule $A$, following $P(A\mid\mathbf{x},\mathcal M_{\mathrm{ret}};\Theta)$. The schedule always includes the \textit{model auxiliary tool} and \textit{exogenous analysis tool}; other diagnostics are selected for the current sequence. Retrieved schedules provide procedural context for choosing these optional tools, while their observations indicate what evidence supported the \mbox{associated forecasts}.

On the first attempt, planning uses the current input and retrieved experiences. If a candidate forecast fails format or logic validation, reflection supplies natural-language feedback for a subsequent planning attempt within the bounded refinement loop. The planner selects evidence sources, and the action module executes the selected tools deterministically to return the ensemble prior and diagnostic evidence. Numerical prediction and forecast validation are described in the \mbox{following subsection}.

\paragraph{Multi-View Evidence Acquisition}
\label{sec:toolkit}
The toolkit comprises eleven tools in four functional categories; Appendix~\ref{app:tool_details} provides the per-tool inventory and execution scopes (Table~\ref{tab:evidence_tools}). The foundational anchorer returns an ensemble prior from forecasting models selected using historical performance on related windows; its retrieval and aggregation are detailed in Section~\ref{sec:forecast_refinement}. The remaining categories characterize the observed sequence from \mbox{complementary perspectives}.

The main inference protocol supplies both the observed lookback and covariate forecasts for the prediction horizon that were issued no later than the forecast origin $t$. We denote the latter by $\mathbf u^{[t]}$. These supplied forecasts are held fixed for the request, including its reflection retries; later-issued revisions and realized future observations are excluded. The execution interface makes this information available to relevant tools, while each tool consumes the inputs supported by its function. In particular, lookback diagnostics do not require future covariate forecasts. Appendix~\ref{app:reproducibility} separates the input protocol, per-value admissibility, and \mbox{dataset provenance}.

The statistical and spectral profiler summarizes numerical scale and dispersion, including robust deviation measures, and uses spectral entropy to describe how power is distributed across frequencies. Its data-quality outputs quantify missingness and provide range bounds for interpreting candidate forecasts. The dynamics monitor examines trends, temporal differences, and lead--lag or cross-channel associations. Its event summary expresses dominant patterns as rise, fall, flat, or oscillation, providing qualitative context alongside \mbox{numerical features}.

The residual diagnoser fits an autoregressive proxy to the observed lookback and summarizes its residual mean and first-order autocorrelation. It runs only during training and is bypassed at inference. Its lookback-fitted residuals do not require future targets, and any such outputs retained in memory describe their historical training window. Standard diagnostic definitions are collected in Appendix~\ref{app:tool_details}.

\subsection{Evidence-Grounded Forecast Refinement}
\label{sec:forecast_refinement}
\label{sec:reasoning}
Tool reasoning contributes to forecasting through two complementary inputs: a numerical forecasting prior and diagnostics of the observed temporal structure. The specialized model uses these inputs to generate a refined forecast, while the frozen reflection model checks response validity and consistency with the \mbox{available evidence}.

\paragraph{Ensemble Forecasting Prior}
\label{sec:ensemble_prior}
The foundational anchorer provides a numerical starting point through the mandatory model auxiliary tool. Its historical case library is constructed offline by partitioning past time series into sliding windows and applying K-medoids clustering~\cite{kaufman1990finding}. Each cluster stores a medoid and historical model-performance information for a pool spanning statistical methods, deep learning forecasters, and pretrained time series foundation models. This library serves model selection and weighting; it is distinct from the Experience Memory of procedural tool-use trajectories. Clustered case-based model references and LLM forecast refinement are also used by AlphaCast~\cite{alphacast}; CastFlow supplies the ensemble as a mandatory tool output and trains the specialized forecasting response through \mbox{SFT and RLVR}.

For an input $\mathbf x$, the anchorer retrieves the closest cluster $\mathcal C^*$ using a distance that combines dynamic time warping~\cite{sakoe1978dynamic}, Euclidean distance, and cosine similarity on z-score-normalized representations. Let $\mathcal I_{\mathcal C^*}$ index the associated forecasting models. The ensemble prior is
\begin{equation}
\begin{aligned}
w_k &= \frac{\exp(-\mathcal L_k)}
 {\sum_{q\in\mathcal I_{\mathcal C^*}}\exp(-\mathcal L_q)},\\
\hat{\mathbf y}_{\mathrm{base}} & =
 \sum_{k\in\mathcal I_{\mathcal C^*}} w_k f_k(\mathbf x),
\end{aligned}
\label{eq:ensemble_prior}
\end{equation}
where $f_k$ is a forecasting expert and $\mathcal L_k$ is its historical validation loss used to derive weights for the retrieved cluster. These historical quantities are separate from the unknown targets of the current forecasting request. The ensemble supplies an initial estimate of future values; the diagnostic tools supply context for interpreting and \mbox{revising that estimate}.

The expert pool and its historical performance records are prepared offline. This pool is a separate configuration from the competing baselines; Appendix~\ref{app:reproducibility} distinguishes expert weighting, memory retrieval, and \mbox{response-generation settings}.

\paragraph{Specialized Temporal Reasoning}
\label{sec:specialized_reasoning}
For each round, the action stage constructs the core forecasting context
\begin{equation}
\mathcal C_j^{\mathrm{core}} =
 \bigl(\mathbf x,\mathbf u^{[t]},\hat{\mathbf y}_{\mathrm{base},j},\mathcal D_j\bigr).
\label{eq:forecast_context}
\end{equation}
The historical observations describe the measured sequence and historical covariates; $\mathbf u^{[t]}$ supplies ex ante auxiliary forecasts aligned with the prediction horizon. Diagnostic outputs characterize distributional and temporal patterns, and the ensemble prior supplies a numerical reference. The full context $\mathcal C_j$ comprises $\mathcal C_j^{\mathrm{core}}$ together with retrieved procedural guidance and available revision feedback. The forecasting prompt can therefore use horizon-specific auxiliary forecasts without receiving the unknown future target or realized future \mbox{covariate observations}.

The specialized LLM generates a serialized response $r_j$, intended to contain a reasoning trace $z_j$ and a numerical forecast:
\begin{equation}
\begin{aligned}
r_j&\sim \pi_\theta(\,\cdot\mid\mathcal C_j),\\
\hat{\mathbf y}_j&=\operatorname{Parse}(r_j).
\end{aligned}
\label{eq:evidence_conditioned_generation}
\end{equation}
The parser returns a finite numerical sequence of the required shape for a structurally valid response and a failure symbol $\perp$ otherwise. The prior supplies a value at each forecast position, while trends, changes, and lagged associations provide context for the direction and timing of possible revisions. Origin-issued covariate forecasts supply conditions specific to the future interval. SFT examples and the accuracy-based rewards in Section~\ref{sec:optimization} train the policy to map this combined context to a complete numerical forecast. The schematic in Fig.~\ref{fig:case1} illustrates how retrieved experience, diagnostic evidence, and the ensemble prior connect to the specialized forecaster. These inputs condition learned numerical generation; they do not specify deterministic correction coefficients or certify the \mbox{generated reasoning trace}.

Refinement can be analyzed by defining $e_{\mathrm{base}}=\mathbf y-\hat{\mathbf y}_{\mathrm{base}}$ and $\Delta_\theta=\hat{\mathbf y}-\hat{\mathbf y}_{\mathrm{base}}$. Then
\begin{equation}
\begin{aligned}
&\|\hat{\mathbf y}-\mathbf y\|_F^2-
 \|\hat{\mathbf y}_{\mathrm{base}}-\mathbf y\|_F^2\\
&\qquad=\|\Delta_\theta\|_F^2
 -2\langle e_{\mathrm{base}},\Delta_\theta\rangle_F .
\end{aligned}
\label{eq:refinement_error_change}
\end{equation}
The Frobenius norm and inner product sum over the forecast horizon and target channels. A correction improves squared error when its alignment with the baseline residual outweighs its squared magnitude. Here, $\Delta_\theta$ is the difference between two forecasts, not a separate residual-prediction head. The identity motivates evaluating improvement against the prior during training; it does not guarantee that every generated correction improves \mbox{forecasting accuracy}.

\paragraph{Forecast Verification and Bounded Revision}
\label{sec:workflow}
After generation, the workflow combines deterministic format checks with a frozen-model assessment of \mbox{evidence consistency}:
\begin{equation}
v_j=\begin{cases}
0, & \mathbb I_{\mathrm{format}}(r_j)=0,\\
\mathbb I_{\mathrm{logic}}(\hat{\mathbf y}_j,\mathcal D_j),
& \mathbb I_{\mathrm{format}}(r_j)=1.
\end{cases}
\label{eq:forecast_verification}
\end{equation}
The format indicator checks whether the response can be parsed into finite numerical values with the required forecast length. For a structurally valid response, the logic indicator records the reflection model's assessment of its consistency with the diagnostic signals. This assessment has no access to the request's unknown forecasting targets and is not a measurement of \mbox{true predictive error}.

If $v_j=0$, reflection supplies feedback for another planning attempt, which may revise the diagnostics or the forecasting prompt. The workflow permits one initial attempt and at most $C_{\max}$ retries. When the budget is exhausted without acceptance, the main evaluation first uses a structurally valid response from the request history, then falls back to the ensemble prior if no such response is available. Only when neither source provides a usable forecast is the output treated as missing or abnormal. This fallback order uses available request outputs without consulting forecasting targets. A fallback prediction is distinct from a response accepted by reflection. Algorithm~\ref{alg:castflow_inference} summarizes the inference \mbox{and fallback sequence}.

\begin{algorithm}[t]
\caption{Bounded inference \mbox{and fallback}}
\label{alg:castflow_inference}
\small
\begin{algorithmic}[1]
\Require Lookback $\mathbf x$, origin-issued covariate forecasts $\mathbf u^{[t]}$, Memory $\mathcal M$, retry budget $C_{\max}$
\Ensure Numerical forecast $\hat{\mathbf y}_{\mathrm{out}}$ or a failure status
\State $\mathcal M_{\mathrm{ret}}\gets\operatorname{Retrieve}(\mathbf x,\mathcal M)$
\State $\mathcal H\gets\emptyset$; $\mathcal R_{\mathrm{valid}}\gets\emptyset$
\For{$j=0,\ldots,C_{\max}$}
  \State $A\gets\operatorname{Plan}_\Theta(\mathbf x,\mathcal M_{\mathrm{ret}},\mathcal H)$
  \State $(\hat{\mathbf y}_{\mathrm{base}},\mathcal D)\gets\operatorname{Execute}(A,\mathbf x,\mathbf u^{[t]})$
  \State Assemble $\mathcal C$ from inputs, prior, evidence, and feedback
  \State Generate $r\sim\pi_\theta(\cdot\mid\mathcal C)$
  \If{$\mathbb I_{\mathrm{format}}(r)=1$}
    \State Add $r$ to $\mathcal R_{\mathrm{valid}}$
  \EndIf
  \State $v\gets\operatorname{Verify}_\Theta(r,\mathcal D)$
  \If{$v=1$}
    \State \Return $\operatorname{Parse}(r)$ \Comment{Accepted}
  \EndIf
  \State Update $\mathcal H$ with observations, $r$, and feedback
\EndFor
\If{$\mathcal R_{\mathrm{valid}}\ne\emptyset$}
  \State Choose $r_{\mathrm{fb}}\in\mathcal R_{\mathrm{valid}}$
  \State \Return $\operatorname{Parse}(r_{\mathrm{fb}})$ \Comment{Response fallback}
\ElsIf{the ensemble prior is usable}
  \State \Return $\hat{\mathbf y}_{\mathrm{base}}$ \Comment{Prior fallback}
\Else
  \State \Return missing or abnormal output status
\EndIf
\end{algorithmic}
\end{algorithm}

\subsection{Outcome-Aligned Training}
\label{sec:optimization}
Forecasting outcomes supervise three distinct offline operations: selecting experiences for memory, selecting teacher responses for SFT, and scoring generated responses for RLVR. The first two retain useful historical procedures and examples; the third optimizes the specialized forecasting policy. Throughout SFT and RLVR, the planner and reflection model remain frozen. Their workflow supplies the conditioning context, while gradient updates apply exclusively to \mbox{the forecasting model}.

\paragraph{Supervised Fine-Tuning}
After constructing the Experience Memory in Section~\ref{sec:evidence_acquisition}, we incorporate it into the workflow to obtain teacher supervision. For each training instance, a powerful teacher model generates candidate forecasting responses conditioned on retrieved procedural guidance, tool-derived diagnostics, and the ensemble prior. We select a preferred valid response using forecasting error over the target horizon, and extract the forecasting context--response pair $(\mathcal C,r^*)$ from the resulting teacher trajectory. The selected response contains a reasoning trace and numerical answer. Upstream plans and tool outputs form the conditioning context rather than additional \mbox{trainable policies}.

Writing the selected response as tokens $(w_1,\ldots,w_T)$, the forecasting model minimizes
\begin{equation}
\mathcal L_{\mathrm{SFT}}(\theta)=
 -\mathbb E_{(\mathcal C,r^*)}
 \left[\sum_{i=1}^{T}
 \log\pi_\theta(w_i\mid w_{<i},\mathcal C)\right].
\label{eq:sft_objective}
\end{equation}
This stage teaches the local model to follow the forecasting response protocol and condition numerical generation on the workflow evidence. Forecasting targets are used to select teacher examples during offline preparation; they are not included as future observations in \mbox{the inference context}.

\paragraph{Verifiable Forecasting Rewards}
For a training instance with target $\mathbf y$, let $r$ be a sampled forecasting response and $\hat{\mathbf y}(r)$ its final numerical prediction when accepted by the numerical reward parser. Each response is scored directly from this prediction; a separate reflection and retry loop is not executed for each GRPO candidate. Thus, final forecasting outcome refers here to the numerical answer in the response conditioned on workflow evidence. We evaluate its horizon-wide MSE and that of the ensemble prior supplied in the same context:
\begin{equation}
\begin{aligned}
\mathcal L_{\mathrm{agent}} & =
 \frac{1}{H C_y}\sum_{h=1}^{H}
 \|\hat{\mathbf y}_{t+h}(r)-\mathbf y_{t+h}\|_2^2,\\
\mathcal L_{\mathrm{base}} & =
 \frac{1}{H C_y}\sum_{h=1}^{H}
 \|\hat{\mathbf y}_{\mathrm{base},t+h}-\mathbf y_{t+h}\|_2^2.
\end{aligned}
\label{eq:forecast_training_losses}
\end{equation}
Both errors refer to the same instance and horizon. The context and its ensemble prior remain fixed within a GRPO group, so each candidate is compared with the same numerical starting point. The reward combines numerical validity, absolute accuracy, and improvement over that prior:
\begin{equation}
\begin{aligned}
R(r)&=\begin{cases}
-\mathcal P_{\mathrm{violation}}(r), & r\notin\mathcal V,\\
R_{\mathrm{abs}}(\mathcal L_{\mathrm{agent}})
 +R_{\mathrm{rel}}(r), & r\in\mathcal V,
\end{cases}\\
R_{\mathrm{rel}}(r)&=\operatorname{Clip}\!\left(
 \lambda\frac{\mathcal L_{\mathrm{base}}-\mathcal L_{\mathrm{agent}}}{\nu_d},
 -\delta,\delta\right).
\end{aligned}
\label{eq:forecast_reward}
\end{equation}
Here, $\mathcal V$ contains responses accepted by the numerical reward parser with the required forecast length and a non-NaN candidate MSE, and $\mathcal P_{\mathrm{violation}}(r)>0$ penalizes unusable answers or length mismatches. For dataset $d$, the absolute term is
\begin{equation}
R_{\mathrm{abs}}(\epsilon)=
\begin{cases}
1-0.9\sin\!\left(\dfrac{\pi\epsilon}{2\gamma_d}\right),
 & 0\leq\epsilon\leq\gamma_d,\\
0.3\exp\!\left(-\dfrac{\ln(3)\epsilon}{\gamma_d}\right),
 & \epsilon>\gamma_d.
\end{cases}
\label{eq:absolute_forecast_reward}
\end{equation}
The branches meet at $R_{\mathrm{abs}}(\gamma_d)=0.1$. The relative term uses $\lambda=5$ and $\delta=0.5$. Its denominator $\nu_d$ is a configured dataset-specific MSE scale, so it measures a scaled error difference rather than a percentage change relative to baseline MSE. Appendix~\ref{app:reward_configuration} gives all dataset scales \mbox{and output penalties}.

The reward is computed directly from each response using offline targets. Deterministic parsing and error computation provide the verifiable training signal. The reward parser determines numerical validity; the reflection model's semantic decision is not part of $\mathcal V$. Accordingly, a candidate accepted by the reward parser can receive a numerical reward independently of whether reflection would accept it during inference. Reflection and fallback govern deployed inference as described in Section~\ref{sec:forecast_refinement}; they are not additional GRPO scoring stages. The reward encourages accurate refinement without directly verifying the faithfulness of \mbox{a reasoning trace}.

\paragraph{Selective Policy Optimization}
We use group relative policy optimization (GRPO)~\cite{shao2024deepseekmath} to optimize the forecasting response policy without a separate value network. For each workflow-generated forecasting prompt $\mathcal C$, the policy samples a group of $G$ responses while holding the observations, diagnostic outputs, procedural guidance, and ensemble prior fixed. Their rewards are compared within the group; for nonzero reward variance, the normalized advantage is
\begin{equation}
A_i=\frac{R_i-\mu_R}{\sigma_R},
\label{eq:grpo_advantage}
\end{equation}
where $\mu_R$ and $\sigma_R$ are the group reward mean and standard deviation. The outcome-aligned objective can be written as
\begin{equation}
J(\theta)=\mathbb E_{\mathcal C\sim d_{\mathcal W},\,
 r\sim\pi_\theta(\cdot\mid\mathcal C)}[R(r)].
\label{eq:outcome_aligned_objective}
\end{equation}
Here, $d_{\mathcal W}$ denotes the distribution of forecasting contexts supplied by the workflow with frozen planning and reflection, and $R(r)$ scores the numerical answer generated under a given context. The objective expresses candidate-level optimization of evidence-conditioned forecasts. The same scalar outcome reward supplies the advantage for the generated response tokens, including its reasoning and answer; upstream plans and tool outputs remain conditioning inputs. Planning and reflection receive no policy updates. Appendix~\ref{app:workflow_formalization} describes the surrounding decision process, while the trainable objective above concerns the specialized \mbox{forecasting response}.

\section{Experiments}
\label{sec:experiments}

\subsection{Experimental Settings}
\label{sec:experimental_settings}

\paragraph{Datasets}   
\input{Datasets}
Table~\ref{tab:datasets_summary} summarizes the ten benchmarks, target variables, and forecasting configurations.
For short-term forecasting, we use the BE, DE, FR, NP, and PJM markets from the EPF benchmark~\cite{lago2021epf}, distributed in the TimeXer collection~\cite{wang2024timexer}. Each contains a target electricity price series and two market-specific exogenous forecast series.
In the long-term setting, the widely used ETTh and ETTm datasets~\cite{zhou2021informer} are utilized to monitor transformer temperature and load variations under different time granularities.
To capture the dynamics of renewable energy generation, we adopt the wind power (WP) and solar power (SP) datasets~\cite{xfyun_renewable_power_challenge_2025}, which align real power generation records with multi-dimensional meteorological conditions.
Additionally, the MOPEX dataset~\cite{schaake2006us} is included for streamflow forecasting, featuring streamflow series supported by climatic factors. 
For rigorous evaluation, all multivariate time series datasets are strictly split chronologically into training, validation, and testing sets with a consistent ratio of 7:1:2.

\paragraph{Baselines}
To provide a comprehensive evaluation, we compare CastFlow against 21 representative baselines spanning five distinct methodological paradigms, covering both conventional and emerging approaches, from classical forecasting to recent reasoning-oriented frameworks under the dataset and metric protocol described below: (1) statistical models: Prophet~\cite{taylor2018forecasting} and ARIMA~\cite{ARIMA}; (2) machine learning models: XGBoost~\cite{chen2016xgboost} and LightGBM~\cite{ke2017lightgbm}; (3) deep learning forecasters: Autoformer~\cite{wu2021autoformer}, DLinear~\cite{zeng2023transformers}, PatchTST~\cite{nie2022time}, iTransformer~\cite{liu2023itransformer}, TimeXer~\cite{wang2024timexer}, and ConvTimeNet~\cite{cheng2025convtimenet}; \mbox{(4) foundation models: Chronos}~\cite{ansari2024chronos}, TimesFM~\cite{das2023timesfm}, and Sundial~\cite{liu2025sundial}; (5) large language model (LLM)-based and agentic frameworks: Time-LLM~\cite{jin2023time}, PromptCast~\cite{xue2023promptcast}, TokenCast~\cite{tao2025tokencast}, S$^2$IP-LLM~\cite{pan2024s2ip}, TimeReasoner~\cite{timereasoner}, Time-R1~\cite{time_r1}, TimeSeriesScientist~\cite{tsscientist}, and AlphaCast~\cite{alphacast}. These baselines provide a solid basis for evaluating our method across \mbox{diverse benchmarks}.

\paragraph{Implementation Details}
We utilize Grok 4~\cite{xai2025grok4} as the frozen backbone model for general-purpose reasoning during both training and testing phases, which also facilitates experience generation as a teacher model during memory construction. Meanwhile, Qwen3-4B~\cite{yang2025qwen3technicalreport} is employed specifically as the trainable local LLM for specialized numerical forecasting, configured with a max completion length of 5,000 tokens. Local model experiments use two NVIDIA A800 GPUs; Grok 4 provides the frozen general-purpose reasoning and teacher responses. We implement the training pipeline using the transformers Trainer~\cite{wolf2020transformers} for the supervised stage and the Agent Lightning framework~\cite{luo2025agentlightning} for the reinforcement learning with verifiable rewards (RLVR) stage. The process consists of two phases: (1) supervised fine-tuning (SFT) with a learning rate of $5 \times 10^{-5}$ and batch size of 8, running for 1 epoch in cross-domain joint training; and (2) RLVR using group relative policy optimization (GRPO)~\cite{shao2024deepseekmath} with a group size of $G=8$, temperature of 1.0, and learning rate of $2 \times 10^{-6}$. The KL penalty coefficient is set to $\beta=0.0$. The RLVR stage spans 3 epochs in \mbox{cross-domain joint training}.

For evaluation, we set the lookback window $L=168$ and horizon $H=24$ for short-term tasks, while long-term tasks use $L=96$ and $H=96$. All methods are evaluated using the same chronological splits, target variables, and forecasting horizons. Methods that support exogenous inputs receive covariates through their supported interfaces; target-only baselines use only the target series. CastFlow also uses the origin-issued horizon covariate forecasts described in Section~\ref{sec:evidence_acquisition}. We follow official implementations or recommended configurations whenever available, including model-specific training schedules, early stopping, preprocessing, and prompt construction. Trainable baselines are fitted on the training split, while remaining hyperparameters and early-stopping criteria are selected on the validation split without test-set information. For prompt- or LLM-based baselines, contextual inputs are constructed according to their original protocols and supported covariate interfaces, restricted to information available at the forecast origin, without realized future observations or test labels. When model-specific preprocessing or normalization is applied, all reported metrics are computed after inverse transformation to the original target scale. CastFlow retains the original magnitude values in its forecasting context without explicit input-output normalization strategies such as RevIN~\cite{kim2021reversible}.

\input{main_table}

Lookback lengths, forecasting horizons, and strides in Table~\ref{tab:datasets_summary} are measured in sampling steps, so their temporal spans depend on dataset-specific sampling intervals, while strides determine the spacing between forecast origins. A 24-step horizon spans 24 hours on BE, DE, FR, NP, and PJM, whereas a 96-step horizon corresponds to 96 hours on ETTh, 24 hours on ETTm, WP, and SP, and 96 \mbox{days on MOPEX}.

\input{Ablation-1}

\paragraph{Generation and ablation reporting.}
The sampling parameters above describe GRPO training. Greedy decoding in the recovery experiment is specific to that execution comparison. Appendix~\ref{sec:result_configuration} clarifies the scope of the reported settings.

\paragraph{Workflow execution support.}
We use request-local diagnostic feature reuse to avoid repeated primitive computations and a persistent execution journal to resume committed stages after interruption. Reuse requires matching inputs and configuration; a semantic revision produces a new forecasting stage. Appendix~\ref{app:execution_support} gives the mechanisms and their scope, and Section~\ref{sec:system_evaluation} evaluates diagnostic computation and controlled \mbox{interruption recovery}.

\subsection{Overall Forecasting Performance}
\label{sec:main_results}
Table~\ref{tab:main_results} compares CastFlow with 21 baselines using MSE and MAE on ten datasets. CastFlow achieves the lowest values of both metrics on nine datasets, corresponding to 18 of the 20 dataset--metric comparisons. It leads on four of the five electricity-market datasets and all five datasets evaluated at the 96-step horizon. These results establish the effectiveness of the complete forecasting configuration across \mbox{the evaluated tasks}.

The gains vary by dataset and metric. On WP, MSE decreases from the strongest competing result of 1932.59 to 1719.99, a reduction of 11.00\%; MAE decreases from 28.52 to 27.40, a reduction of 3.93\%. On MOPEX, the corresponding reductions are 25.47\% in MSE, from 4.83 to 3.60, and 7.03\% in MAE, from 1.28 to 1.19. Each percentage uses the lowest competing value for the same dataset and metric as its reference. Improvements are smaller on several other comparisons, including BE MAE (9.63 to 9.60) and ETTh MSE (8.02 to 8.00). The table reports point estimates, so these small margins should not be interpreted as evidence of statistically \mbox{significant superiority}.

PJM is the exception: CastFlow obtains an MSE of 27.45 and an MAE of 3.74, compared with the best competing values of 25.70 and 3.59, respectively. The complete workflow therefore does not improve on the strongest baseline in every setting. The following analyses examine its components and training under the reported configurations; the aggregate comparison alone does not attribute the gains to a particular reasoning step or explain \mbox{the PJM gap}.

\subsection{Experience-Guided Evidence Acquisition}

We examine Experience Memory, trajectory organization and retrieval, and toolkit composition. The training-built library remains fixed during testing; test responses and targets are never written back. Diagnostics use origin-available inputs, including previously issued horizon covariate forecasts where supported (Appendix~\ref{app:reproducibility}).

\paragraph{Experience Memory Contribution}
The full configuration has lower MSE and MAE than the no-memory variant on all ten datasets (Table~\ref{tab:ablation_full}). Removing memory raises MOPEX MSE from 3.60 to 4.63 and MAE from 1.19 to 1.38, and ETTh MSE from 8.00 to 8.45. Relative to the no-memory variant, MOPEX improves by 22.25\% MSE and 13.77\% MAE. BE gains are smaller: MSE changes from 551.54 to 546.87 and MAE from 9.62 to 9.60, improvements of 0.85\% and 0.21\%. Memory utility is therefore task-dependent. Because guidance enters both planning and forecasting, this ablation measures their combined workflow contribution rather than \mbox{tool selection alone}.

\paragraph{Memory Organization and Retrieval}
\begin{figure*}
    \centering
    \includegraphics[width=1\linewidth]{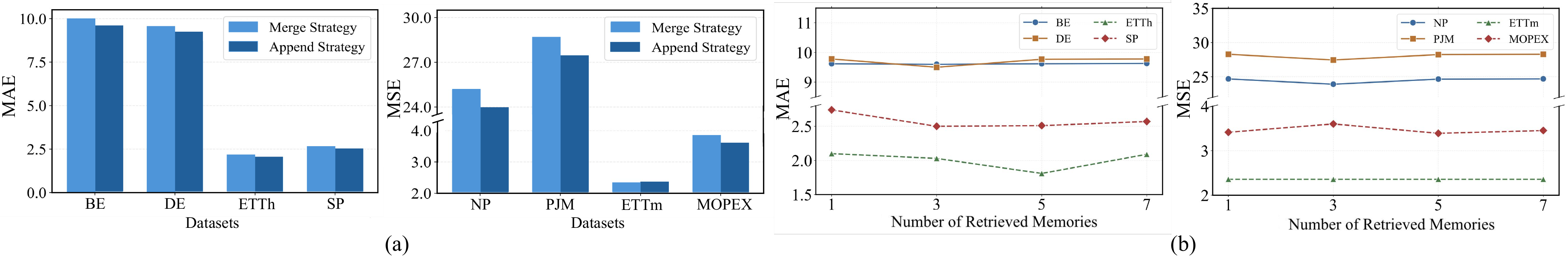}
\caption{Organization and retrieval of training experience. (a) Append and merge are compared before the Experience Memory is fixed for testing. (b) Forecasting error as the retrieval scale $K$ varies}
    \label{fig:exp_bc}
\end{figure*}

\input{toolkit_ablation}

Before the library is fixed, append retains trajectories separately while merge combines them with representative experiences. Append generally yields lower errors, particularly on BE and DE (Fig.~\hyperref[fig:exp_bc]{\ref*{fig:exp_bc}(a)}). Preserving individual tool schedules and diagnostic contexts offers a \mbox{plausible explanation}.

For retrieval scales $K\in\{1,3,5,7\}$, BE and SP show comparatively stable MAE; other datasets improve before plateauing or deteriorating (Fig.~\hyperref[fig:exp_bc]{\ref*{fig:exp_bc}(b)}). More context is therefore not uniformly beneficial, and a small relevant collection can suffice \mbox{in some settings}.

\begin{figure}
    \centering
    \includegraphics[width=1\linewidth]{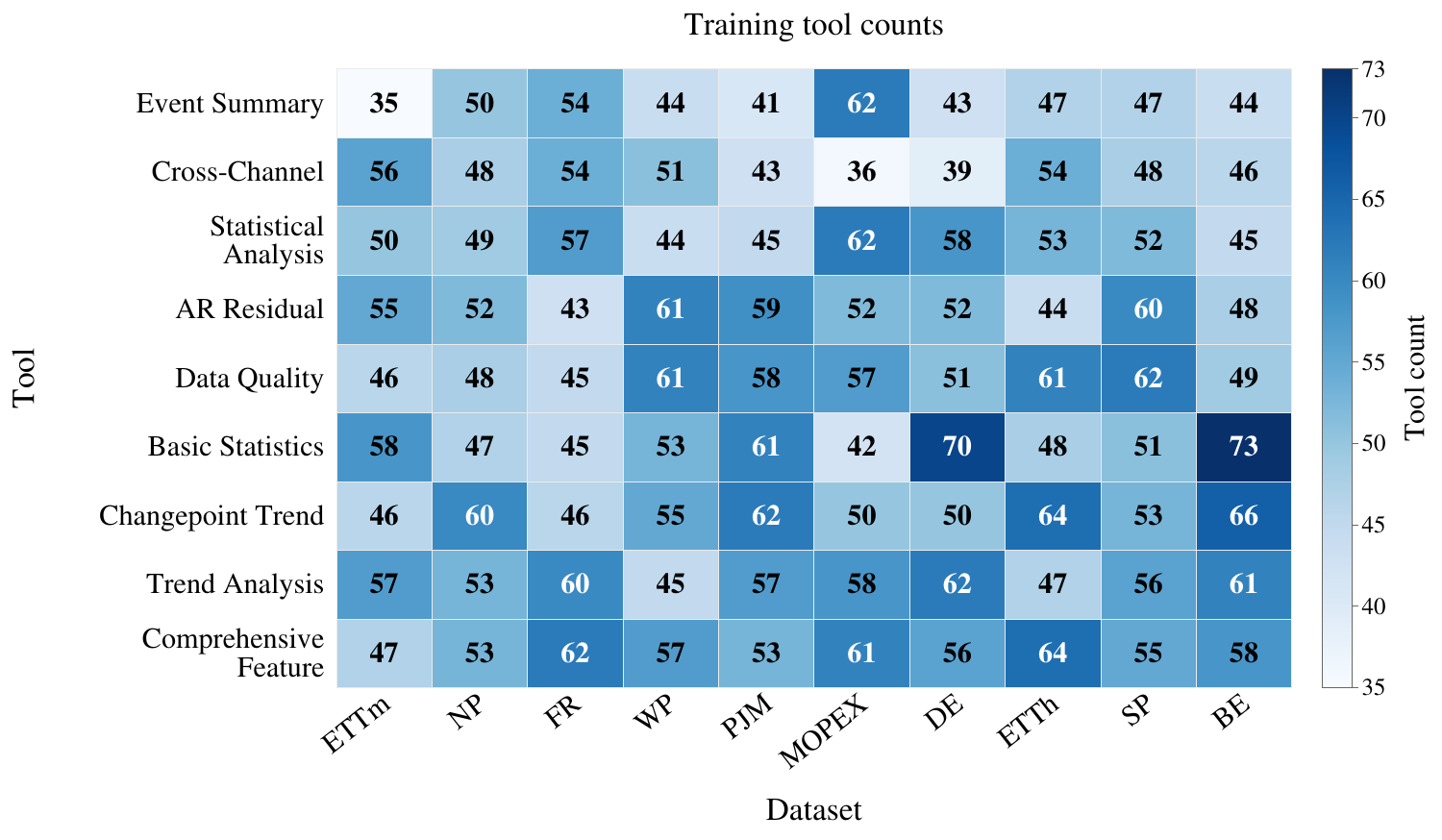}
\caption{Training-phase tool invocation counts across datasets. The two mandatory tools, model auxiliary and exogenous analysis, are omitted. The color scale covers all displayed counts (35--73). Counts describe schedules generated by the \mbox{frozen planner}}
    \label{fig:case}
\end{figure}

\begin{figure}[t]
    \centering
    \includegraphics[width=1\linewidth]{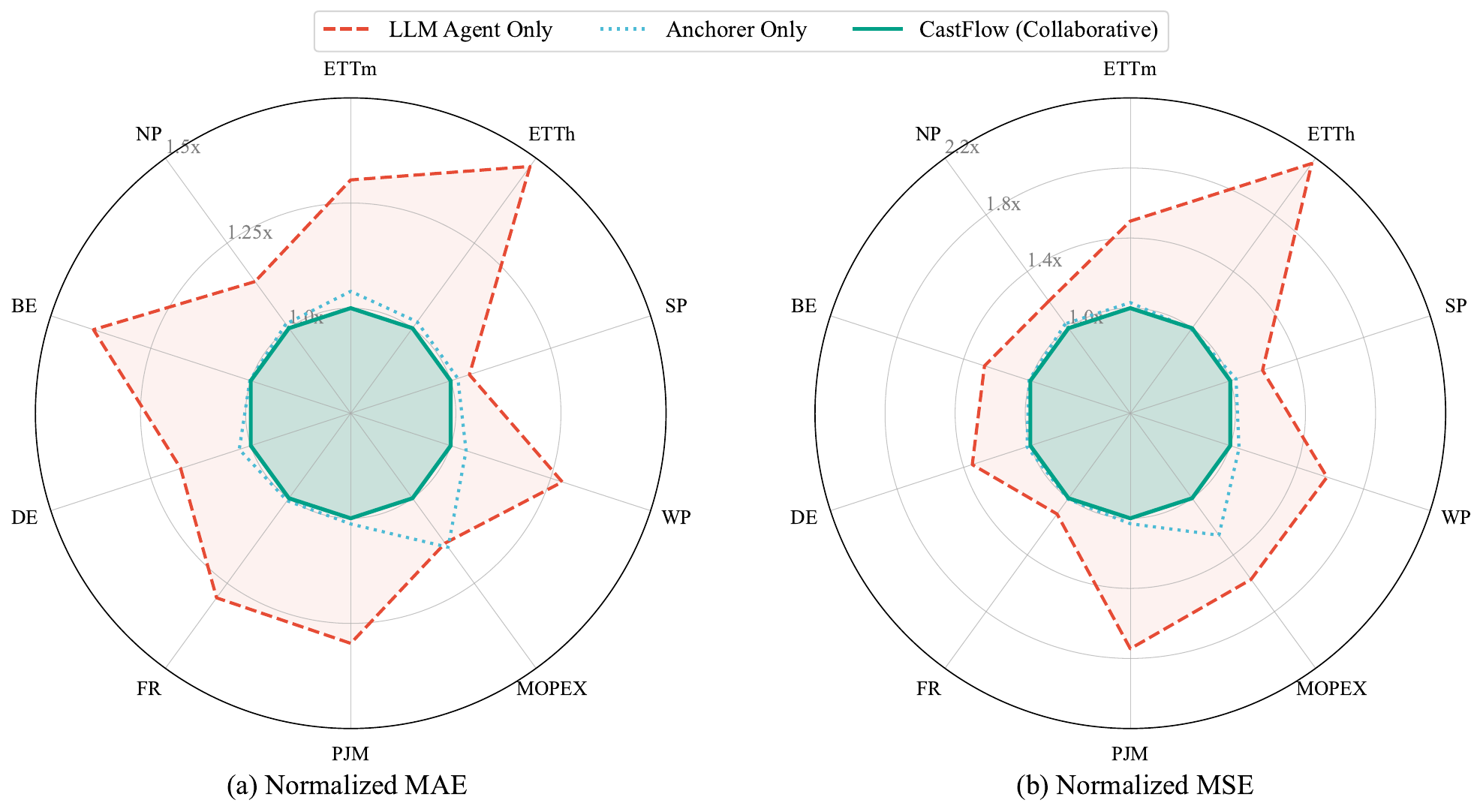}
    \caption{Forecasting configurations relative to full CastFlow: (a) MAE and (b) MSE. LLM Agent Only uses Grok 4 with diagnostics but without an ensemble prior; Anchorer Only returns the prior; CastFlow adds the complete evidence-conditioned \mbox{refinement workflow}}
    \label{fig:collab_radar_dual}
\end{figure}

\paragraph{Tool-Category Contributions}
Removing the anchorer produces the largest MSE increase among category removals on every dataset, including WP's increase from 1719.99 to 2718.42 (Table~\ref{tab:toolkit_ablation}). This pattern confirms the baseline's central role in evidence-guided refinement. Other effects vary: removing the profiler improves FR MSE from 707.96 to 675.67 and MAE from 8.16 to 7.59, reductions of 4.56\% and 6.99\%, but raises MOPEX MSE from 3.60 to 4.69. The full configuration has the lowest MSE on seven datasets and MAE on eight. Adding diagnostics therefore does not uniformly improve refinement, motivating instance-relevant evidence selection. The residual diagnoser runs only during training; retained summaries describe training windows, not \mbox{current test requests}.

\begin{figure*}[!t]
    \centering
    \figcapgap=6pt\relax
    \includegraphics[width=1\linewidth]{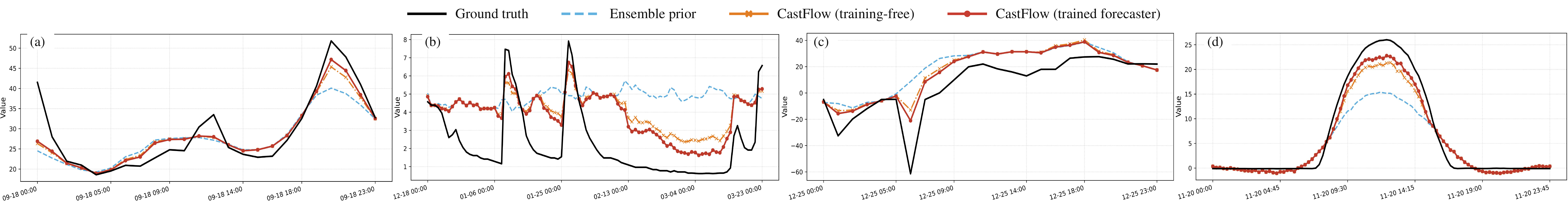}
    \caption{Illustrative forecasting windows (a)--(d), comparing the ensemble prior, training-free CastFlow, and CastFlow with the trained forecasting model. Panel letters identify displayed examples; the original curves \mbox{are retained}}
    \label{fig:improve_sample}
\end{figure*}

\begin{figure*}
    \centering
    \includegraphics[width=1\linewidth]{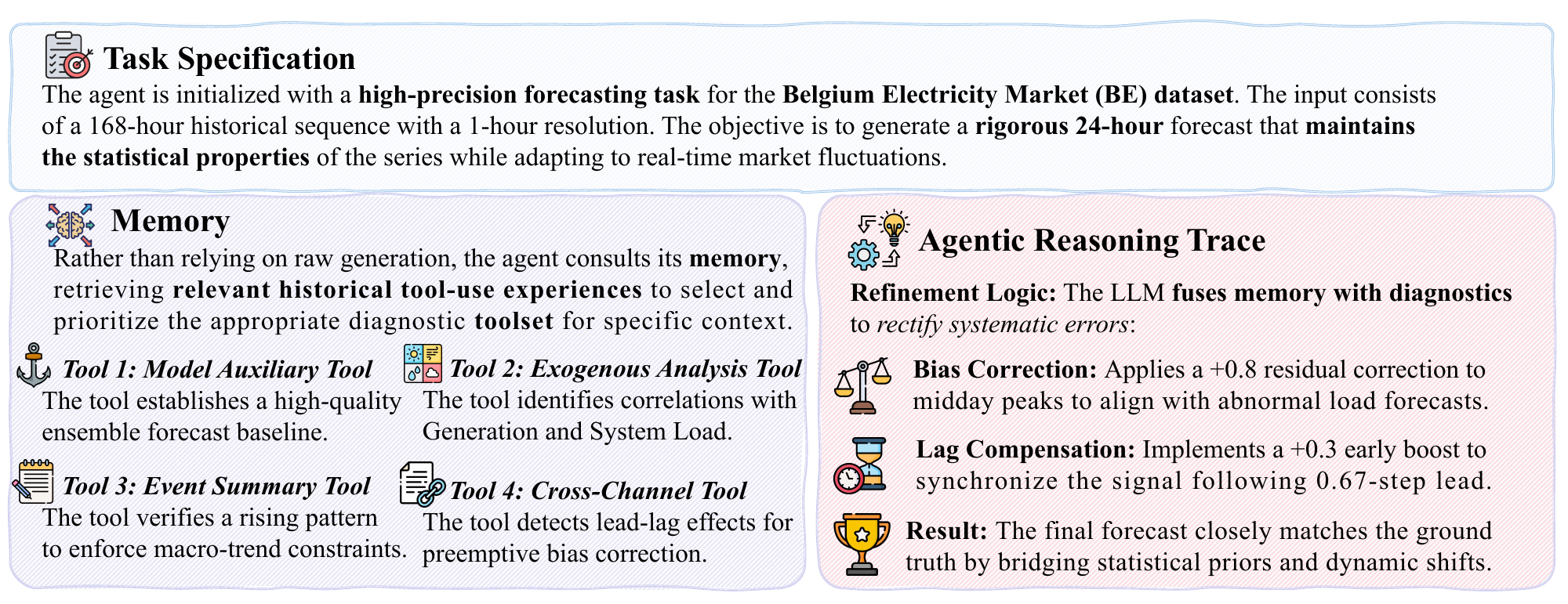}
\caption{Illustrative BE forecasting workflow with a 168-hour lookback and a 24-hour horizon. Retrieved experience guides tool planning; the ensemble prior and diagnostic evidence condition the forecasting response. The displayed trace describes peak and \mbox{timing adjustments}}
    \label{fig:case1}
\end{figure*}

\paragraph{Observed Tool Use}
Figure~\ref{fig:case} shows nine optional tools, excluding mandatory model auxiliary and exogenous analysis. Statistical and trend tools are frequent, particularly on BE and DE; event summaries are less frequent. These training schedules describe the frozen planner, whose tool-selection policy is not directly updated by \mbox{forecasting-model training}.

\subsection{Evidence-Grounded Forecast Refinement}

\paragraph{Forecasting Configurations}
Figure~\ref{fig:collab_radar_dual} compares Grok 4 with diagnostics but no prior (LLM Agent Only), the ensemble prior (Anchorer Only), and full CastFlow. The first generally has larger normalized errors; the prior is competitive, and full refinement improves the overall profile. Both prior availability and forecasting-model adaptation differ, so this comparison does not isolate role \mbox{specialization alone}.

\paragraph{Toolkit Support and Response Verification}
Removing the toolkit raises WP MSE from 1719.99 to 3517.03 and ETTh MSE from 8.00 to 23.71 (Table~\ref{tab:ablation_full}); this removes the prior, diagnostics, and dependent planning together. Removing reflection yields WP MSE of 5102.29; this result does not isolate semantic reflection from structural checks or terminal output handling. Neither comparison isolates diagnostic reasoning or semantic reflection alone. Appendix~\ref{app:training_output_protocol} specifies the fallback order used in the main evaluation when no response passes the verification checks within the retry budget.

\paragraph{Illustrative Forecast Refinement}
Figure~\ref{fig:improve_sample} illustrates how refinement changes the ensemble prior in four windows. In panels (a) and (d), the trained forecaster's curve more closely follows the positive peak than the ensemble prior. The benefits have limits: the abrupt negative excursion in panel (c) remains poorly captured by both refinement configurations. These examples show changes in peak magnitude and local alignment together with remaining errors. They complement the forecasting tables as qualitative illustrations, without establishing population-level improvement in peak \mbox{or lag correction}.

The BE example in Fig.~\ref{fig:case1} connects refinement to diagnostic context for a 168-hour lookback and a 24-hour horizon. Retrieved experience guides the frozen planner's tool schedule. The model auxiliary tool establishes the ensemble prior; exogenous analysis, event summaries, and cross-channel associations provide context for the specialized forecasting response. The displayed narrative describes peak and timing adjustments, illustrating how a numerical prior and temporal diagnostics can be combined in a forecasting explanation. This example concerns the intended information flow and a selected forecast; it does not independently establish the faithfulness of the \mbox{generated rationale}.

\subsection{Outcome-Aligned Training}

\paragraph{Training Stages}
\begin{figure}[!t]
    \centering
    \includegraphics[width=1\linewidth]{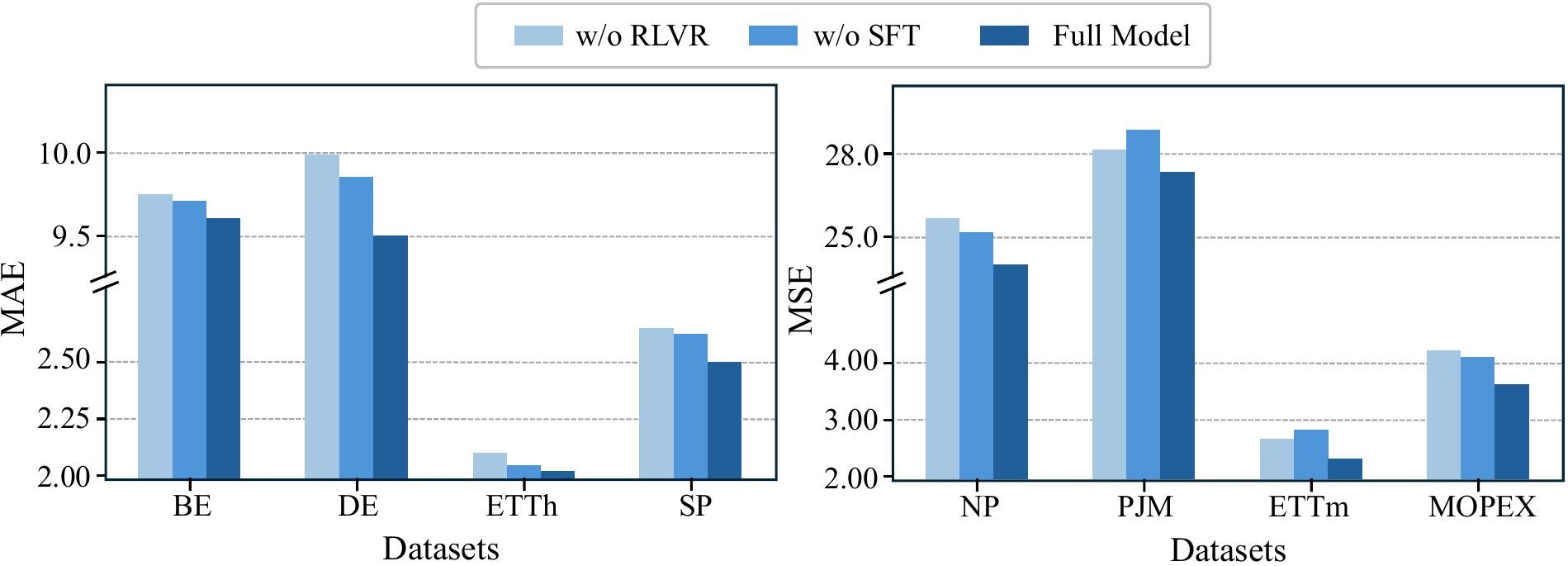}
    \caption{Training-stage ablation for the specialized forecasting model. The full SFT followed by RLVR configuration is compared with variants omitting either stage using MAE and MSE; the planning and reflection model \mbox{remains frozen}}
    \label{fig:ab_rl}   
\end{figure}

We examine how the two training stages contribute to specialized forecast refinement. Figure~\ref{fig:ab_rl} compares the full model with variants omitting SFT or RLVR. The full model achieves the lowest error for every plotted dataset--metric pair, covering MAE on BE, DE, ETTh, and SP, and MSE on NP, PJM, ETTm, and MOPEX. Removing either stage increases error, supporting the combined use of supervised initialization and forecasting-outcome optimization in \mbox{the evaluated setting}.

These stages act on the forecasting LLM. SFT provides examples of generating structured forecasting responses from the workflow context, while RLVR evaluates generated forecasts using numerical rewards. Their complementary effects are consistent with this division of training objectives: imitation supplies an initial response policy, and reward optimization further improves the resulting predictions. The planner and reflection model remain frozen, so these results concern adaptation of evidence-conditioned forecasting rather than \mbox{learned tool selection}.

\paragraph{Forecasting Model Adaptation}
\begin{figure}
    \centering
    \includegraphics[width=1\linewidth]{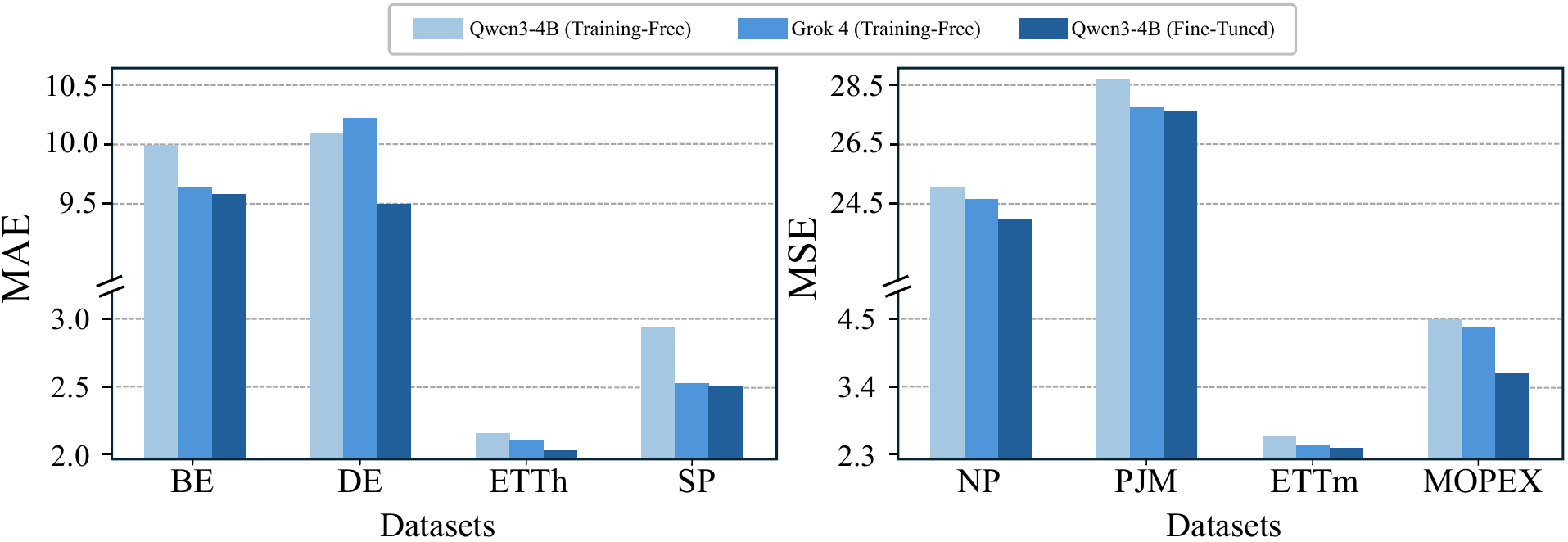}
\caption{Forecasting performance of training-free Grok 4, training-free Qwen3-4B, and trained Qwen3-4B in the evaluated configurations. Training adapts the forecasting model; Grok 4 remains the frozen general-purpose model \mbox{in CastFlow}}
    \label{fig:exp_a}
\end{figure}

Figure~\ref{fig:exp_a} compares the reported training-free Qwen3-4B and Grok~4 configurations with Qwen3-4B after SFT and RLVR. Grok~4 achieves lower error than training-free Qwen3-4B on every plotted comparison except DE, indicating stronger training-free forecasting performance in most of the evaluated settings. After specialized training, Qwen3-4B achieves lower error than both training-free configurations for every plotted dataset--metric pair. The pattern appears across electricity-price, renewable-generation, and streamflow tasks, rather than being confined \mbox{to one application}.

The results support specialized adaptation as a practical route to improving the evaluated forecasting configurations. They also distinguish the numerical forecasting role from the general-purpose reasoning role: a compact model can produce more accurate forecasts after task-specific training, while the frozen Grok~4 backbone continues to support planning and reflection in CastFlow. We interpret this figure as a configuration-level comparison; the training-stage ablation provides the more direct comparison of retaining or removing \mbox{SFT and RLVR}.

\paragraph{Reward Formulation}
\input{reward}
Table~\ref{tab:reward_ablation_final} compares absolute MSE, relative MSE, hybrid MSE, and absolute MAE rewards. The hybrid MSE formulation achieves the lowest MSE on all four reported datasets and the lowest MAE on BE, SP, and ETTh. On DE, absolute MSE yields a slightly lower MAE, 9.49 versus 9.50, while hybrid MSE gives the lowest MSE, 200.47. The benefit is therefore broad across the reported comparisons, with a small \mbox{metric-specific exception}.

The two terms evaluate complementary aspects of the scored forecasting output. The absolute term favors accurate predictions over the full horizon; the relative term evaluates improvement against the ensemble prior for the same instance. Their combination aligns the specialized forecaster with both final prediction quality and the utility of its refinement. Neither term alone matches the hybrid formulation across the reported results. Absolute MAE also produces higher MSE on each of the four datasets. These observations support the selected reward for forecasting refinement under the evaluated protocol, where offline targets score the numerical output and only the specialized model \mbox{receives policy updates}.

\paragraph{Optimization Dynamics}
\begin{figure}
    \centering
    \includegraphics[width=1\linewidth]{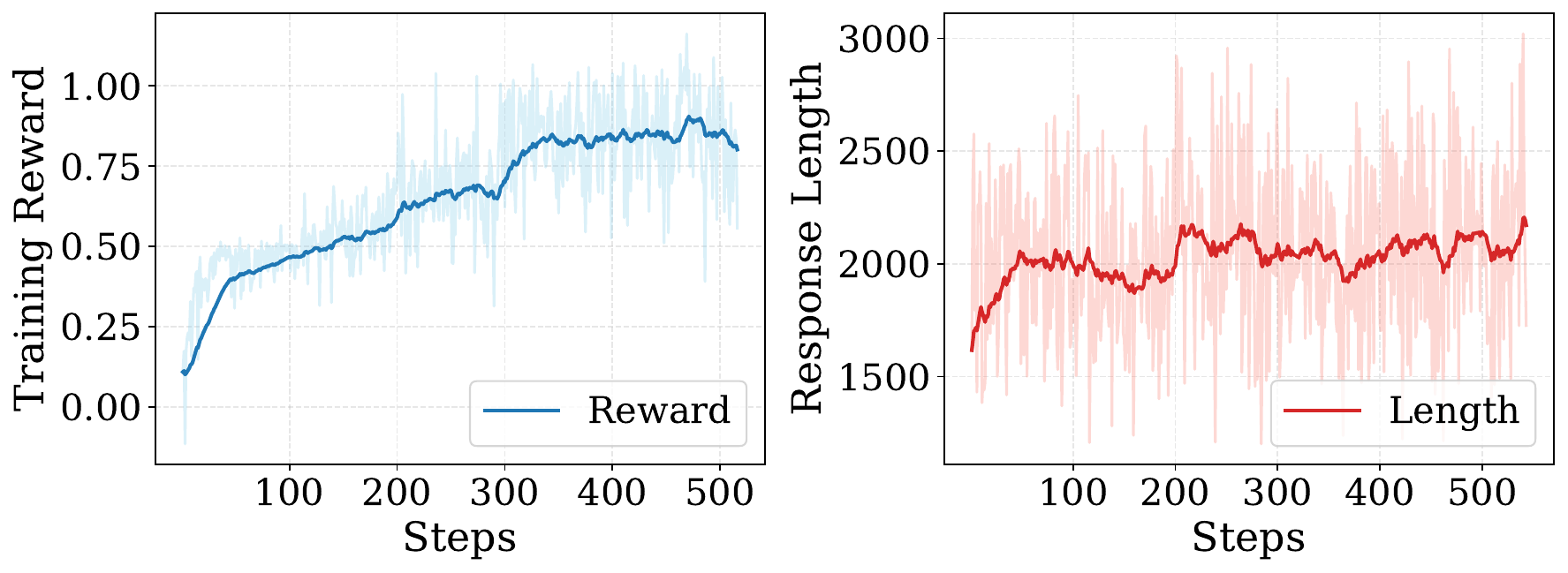}
\caption{GRPO training dynamics of the specialized forecaster. The reward reflects numerical accuracy and output validity; average response length records generated \mbox{text length}}
    \label{fig:training}
\end{figure}

Figure~\ref{fig:training} shows the local model's GRPO training reward and response length. Reward rises rapidly early in training and subsequently increases more gradually, with fluctuations around the later, higher level. Average response length initially grows and then fluctuates within a comparatively stable range. Thus, the observed reward improvement does not require a continuing increase in \mbox{generated text length}.

Because the reward incorporates structural validity as well as numerical forecasting performance, the curve describes progress under the composite training objective. The stage ablation and forecasting comparisons provide complementary evidence about prediction accuracy. Response length characterizes \mbox{the generated output}.

\subsection{Forecasting Backbone Analysis}
\begin{figure}[t]
    \centering
    \includegraphics[width=1\linewidth]{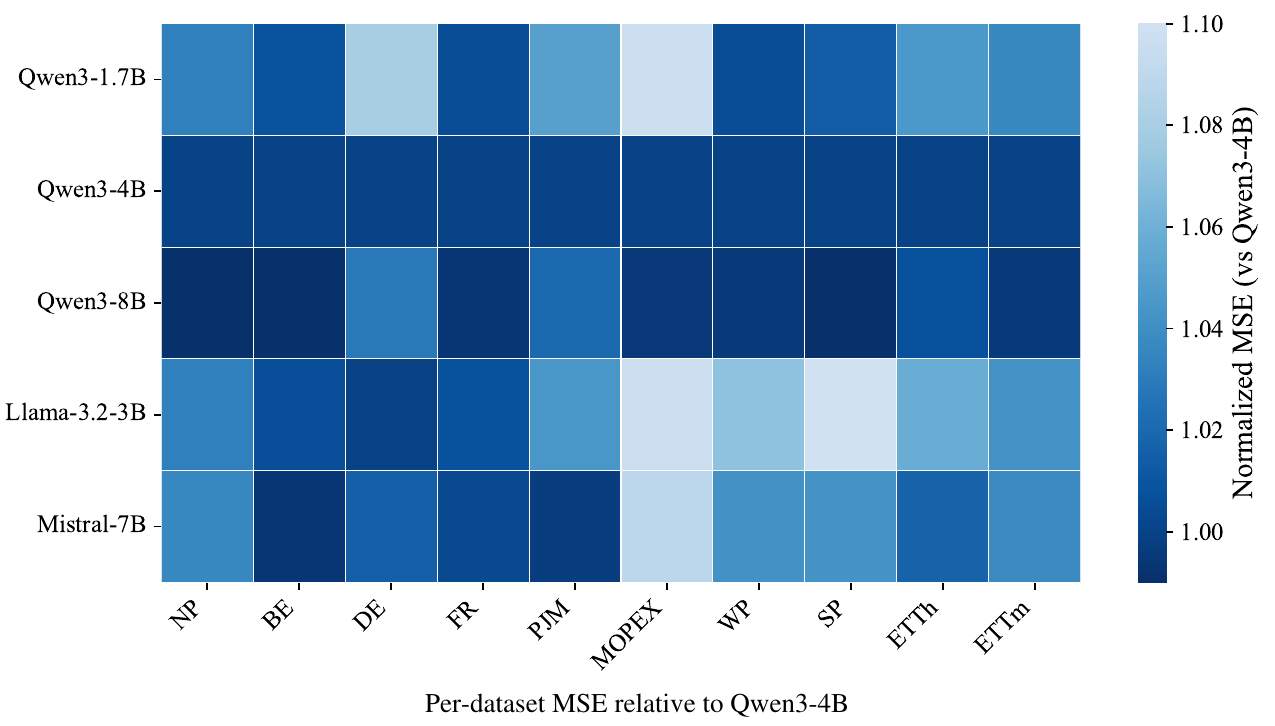}
    \caption{Per-dataset forecasting backbone comparison with the frozen Grok 4 planner retained. Each cell reports MSE relative to Qwen3-4B on the same dataset; a ratio below one indicates lower error. The source per-dataset heatmap is retained without \mbox{cross-dataset aggregation}}
    \label{fig:base_model}
\end{figure}

We compare specialized forecasting backbones while retaining Grok~4 as the frozen reasoning model. Figure~\ref{fig:base_model} shows per-dataset MSE ratios for Qwen3-1.7B, Qwen3-4B, Qwen3-8B, Llama-3.2-3B~\cite{meta2024llama32}, and Mistral-7B~\cite{jiang2023mistral}. For backbone $m$ and dataset $d$, the normalization is $r_{m,d}=\mathrm{MSE}_{m,d}/\mathrm{MSE}_{\mathrm{Qwen3\mbox{-}4B},d}$ within this comparison. The reference row is therefore one, and lower ratios indicate better forecasting performance on that dataset. We interpret these ratios separately \mbox{for each dataset}.

The Qwen3-1.7B variant generally has higher error than the 4B reference, while increasing capacity to 8B does not improve every dataset: the 8B variant has higher relative MSE on DE and ETTh. Llama-3.2-3B and Mistral-7B are competitive on individual datasets, with relative errors varying across domains. Qwen3-4B remains a practical specialized forecaster because it is competitive across the displayed tasks while using fewer parameters than the 8B variant. This comparison supports the selected backbone under the evaluated configurations without requiring an aggregate error across datasets with \mbox{different target scales}.

\subsection{System Efficiency and Execution Recovery}
\label{sec:system_evaluation}
We evaluate the computational efficiency of diagnostic feature reuse and persistent stage recovery.
The two experiments examine repeated feature computation within an action plan and forecast regeneration after interruption, respectively. They measure execution optimizations separately from the \mbox{forecasting comparisons above}.

\paragraph{Diagnostic Feature Reuse}
\begin{figure*}[t]
    \centering
    \includegraphics[width=\textwidth]{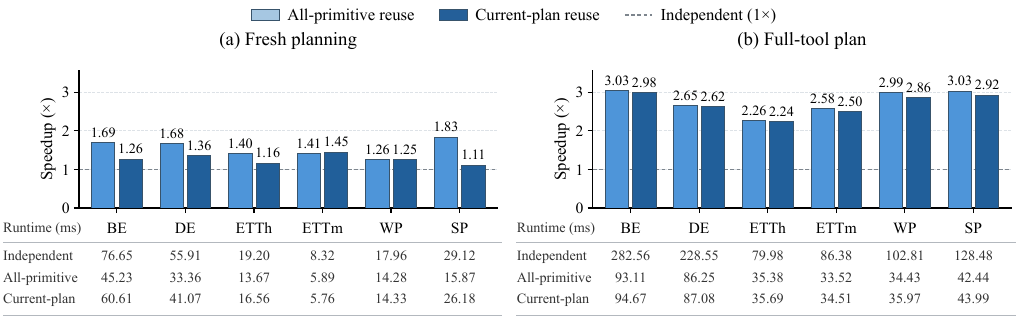}
    \caption{Diagnostic feature reuse under (a) fresh planning and (b) a full-tool plan. Bars show speedup relative to independent computation (dashed line). Aligned rows report runtime in milliseconds, averaged over ten windows and three repetitions. All implementations produce identical \mbox{diagnostic outputs}}
    \label{fig:system_features}
\end{figure*}

To evaluate the computational benefits of diagnostic feature reuse, we compare independent computation, all-primitive reuse, and current-plan reuse under fresh planning and a fixed full-tool plan.
The evaluation uses ten fixed windows per dataset, separate from those used to develop the caching policy, with identical inputs and tool sequences across implementations to isolate the computational effects of feature reuse.
Runtime is averaged over three repetitions after warm-up, with rotated execution order on a single CPU thread.
Measurements cover diagnostic computation only and include all recorded plans, even those that invoke none of \mbox{the shared tools}.

As illustrated in Fig.~\ref{fig:system_features}, both reuse policies reduce mean diagnostic runtime across the evaluated datasets while preserving identical outputs.
Specifically, current-plan reuse achieves speedups of 2.24--2.98 under full-tool plans and 1.11--1.45 under fresh planning.
The larger gains under full-tool plans reflect greater overlap among diagnostic computations that would otherwise be repeated across tools.
All-primitive reuse is generally faster, whereas current-plan reuse reduces the maximum serialized cache payload from approximately 14.5\,kB to 8.9\,kB.
These findings demonstrate that current-plan reuse retains substantial computational savings while limiting the storage of \mbox{intermediate features}.

\paragraph{Recovery after a Controlled Interruption}
\input{system_recovery}
To evaluate the effectiveness of persistent stage recovery, we compare restarting a request with resuming its committed stages on six fixed windows from the training split of each of six datasets (BE, DE, ETTh, ETTm, WP, and SP).
Each pair uses identical inputs, a fixed full-tool plan, deterministic validation, and the same historical ensemble and dataset-specific Qwen3-4B checkpoint trained with SFT and RLVR.
These checkpoints differ from the jointly trained model used in the main forecasting experiment.
We consider normal execution and an exception injected after forecast commitment but before validation, followed by journal reopening.
This interruption point isolates whether completed forecasting work can be retained when the workflow resumes.
Inference uses greedy decoding with a 5,000-token limit, and executor order alternates across pairs to reduce potential order effects.
Timing includes journal I/O and excludes model-service initialization; all attempts, including four ETTm format-validation failures, remain in the timing \mbox{and call statistics}.

As shown in Table~\ref{tab:system_recovery}, the two executors exhibit similar mean runtimes under normal execution.
Under interruption, stage recovery reduces mean execution time across all six evaluated datasets.
For the 35 pairs that reach the scheduled interruption point, recovery preserves identical final predictions and halves forecast calls, achieving a median paired speedup of 1.98.
These gains arise from resuming validation directly from the committed forecast, avoiding another invocation of the forecasting module.
This benefit is especially relevant when forecast generation accounts for a substantial share of execution time, since retaining its output prevents that cost from being incurred again during request recovery.
These results highlight how persistent execution state supports \mbox{efficient workflow recovery}.

\section{Conclusion}
\begingroup\emergencystretch=1em
We presented CastFlow, a role-specialized framework for forecast-relevant evidence acquisition and numerical refinement. Training-built Experience Memory guides a frozen LLM's analytical tool use, while a specialized forecaster refines an ensemble prior from the acquired diagnostics. SFT and verifiable forecasting rewards adapt only this forecaster. Across ten benchmarks, CastFlow achieves the lowest MSE and MAE among evaluated methods on nine datasets. Component and training analyses support the complete workflow while exposing dataset-dependent diagnostic utility and attribution limits. These findings motivate coordinating evidence acquisition with outcome-aligned numerical refinement in agentic time \mbox{series forecasting}.\par\endgroup

\appendix
\FloatBarrier
\section{Diagnostic Tool Definitions}
\label{app:tool_details}
This appendix collects the diagnostic computations used by the toolkit in Section~\ref{sec:evidence_acquisition}. Their outputs characterize the observed lookback window and available covariates; they do not require the current request's future \mbox{forecasting targets}.

\input{tool_specification}

\subsection{Statistical and Spectral Profiler}
The statistical analysis tool returns the mean $\mu$, standard deviation $\sigma$, and boundary extrema. The basic statistics tool includes robust dispersion measures such as
\begin{equation}
\operatorname{MAD}(\mathbf x)=\operatorname{median}(|x_t-\tilde x|),
\end{equation}
where $\tilde x$ is the sequence median. Spectral entropy characterizes the dispersion of spectral power:
\begin{equation}
S_{\mathrm{spec}}(\mathbf x)=-\sum_{k=1}^{\lfloor L/2\rfloor}P_k\log P_k,
\end{equation}
where
\begin{equation}
P_k=\frac{|\mathcal F(\mathbf x)_k|^2}
{\sum_{q=1}^{\lfloor L/2\rfloor}|\mathcal F(\mathbf x)_q|^2}.
\end{equation}
Here, $\mathcal F$ denotes the discrete Fourier transform. Greater entropy corresponds to a more dispersed spectrum. The data quality tool measures dropout ratios and supplies the range diagnostic $\mathcal B=[\mu-\kappa\sigma,\mu+\kappa\sigma]$. The comprehensive feature tool aggregates these outputs as $\mathcal S_{\mathrm{stat}}=\langle\mu,\sigma,\operatorname{MAD},S_{\mathrm{spec}},\mathcal B\rangle$. These quantities provide descriptive evidence for the \mbox{forecaster and verifier}.

\subsection{Dynamics Monitor}
The trend analysis tool computes the linear slope
\begin{equation}
m=\frac{\sum_t(t-\bar t)(x_t-\bar x)}
{\sum_t(t-\bar t)^2}.
\end{equation}
The changepoint trend tool examines the first and second differences, $\Delta x_t=x_t-x_{t-1}$ and $\Delta^2x_t=\Delta x_t-\Delta x_{t-1}$, as diagnostics of changing local dynamics. The cross-channel and exogenous analysis tools examine lagged associations. For an observed target sequence $x$ and auxiliary sequence $u$, the normalized lagged correlation is
\begin{equation}
\rho_{x,u}(d)=
\frac{\sum_t(x_t-\bar x)(u_{t+d}-\bar u)}
{\sqrt{\sum_t(x_t-\bar x)^2\sum_t(u_{t+d}-\bar u)^2}},
\end{equation}
where $d$ denotes the lag and the sums use the available paired observations. The event summary tool complements these numerical features with a dominant qualitative label in $\{\mathrm{rise},\mathrm{fall},\mathrm{flat},\mathrm{oscillation}\}$.

\subsection{Residual Diagnoser}
The autoregressive residual tool fits a proxy process on the observed lookback window and computes
\begin{equation}
\epsilon_t=x_t-\left(c+\sum_{i=1}^{p}\phi_i x_{t-i}\right),
\end{equation}
where $p$ is selected by information criteria. It summarizes residual bias and first-order dependence as
\begin{equation}
\mu_\epsilon=\frac{1}{L-p}\sum_{t=p+1}^{L}\epsilon_t,\qquad
r_1=\frac{\sum_{t=p+2}^{L}\epsilon_t\epsilon_{t-1}}
{\sum_{t=p+1}^{L}\epsilon_t^2}.
\end{equation}
This tool is available only during training and is bypassed at inference. Its lookback-fitted residuals describe an autoregressive proxy, not the unknown future errors of the \mbox{ensemble forecasting prior}.

\section{Workflow Execution Support}
\label{app:execution_support}
Experience Memory provides historical procedural guidance. The following mechanisms reuse computations or committed stages within the execution of a forecasting request. They support the efficiency and recovery experiments in Section~\ref{sec:system_evaluation}.

\subsection{Diagnostic Feature Reuse}
\label{sec:diagnostic_reuse}
Several diagnostic tools share computations such as statistics, temporal differences, and cross-channel summaries. A request-local feature cache avoids recomputing identical operations. Each key identifies the primitive operation, its input values and array structure, and its extraction parameters, including the seasonal period. A missing result is computed once and a copy is returned on subsequent matching calls. Each request starts with an empty cache; computations involving mutable fitted reference \mbox{profiles are excluded}.

The all-primitive policy caches every eligible primitive. The current-plan policy instead caches primitives referenced more than once by the current tool plan, using dependency counts from tool definitions and updating eligibility when the plan changes. It uses the current plan without inspecting later retry plans or forecasting targets. Both policies preserve tool interfaces and diagnostic values. Their benefit depends on actual overlap: cache lookup and copying can offset savings when a plan contains few \mbox{repeated computations}.

\subsection{Persistent Stage Recovery}
\label{sec:stage_recovery}
Forecast revision changes evidence or prompt context to generate a new response. Execution recovery resumes completed work after an interruption with the same forecasting inputs, including any supplied horizon covariate forecasts. A persistent execution journal records completed planning, tool, forecasting, and validation stages. A request identity binds the forecasting inputs, model configuration, generation settings, and implementation version. Stage records contain an input fingerprint, the output, and an output hash. A completed stage is reused only when the request identity and stage inputs match and integrity checks pass. Thus, a request interrupted after forecast commitment can resume validation without \mbox{regenerating that forecast}.

The journal uses a request lock and atomic file replacement with explicit synchronization. Semantic revisions receive new plan and forecast stage identifiers, while previous responses and rejected outputs remain recorded. Only tools declared reusable retain stage identifiers across revisions with unchanged inputs. New requests never inherit another request's numerical predictions. An external call interrupted before durable commitment may need to be repeated; recovery therefore does not provide exactly-once execution of external services. The reported evaluation isolates diagnostic computation and a controlled post-forecast interruption, rather than arbitrary service failures or \mbox{end-to-end scalability}.

\section{Workflow Decision-Process Interpretation}
\label{app:workflow_formalization}
The collaborative workflow can be described by a decision process $(\mathcal S,\mathcal A,\mathcal P,\mathcal R)$. Its state contains the observed lookback, origin-issued horizon covariate forecasts, current ensemble prior, retrieved experience, and workflow history:
\begin{equation}
s_j=(\mathbf x,\mathbf u^{[t]},\hat{\mathbf y}_{\mathrm{base},j},
\mathcal M_{\mathrm{ret}},\mathcal H_j).
\end{equation}
The prior is empty until tool execution produces it. Planning actions select tools, and forecasting actions produce numerical refinements through token generation. Tool execution and the addition of observations and feedback induce transitions. A full trajectory $\tau=(s_1,a_1,\ldots,s_M,a_M)$ is generated by the combined workflow policy $\Pi_{\Theta,\theta}$, comprising frozen planning and reflection, deterministic execution, and the trainable forecasting response policy $\pi_\theta$. The training objective in Eq.~\ref{eq:outcome_aligned_objective} updates only $\theta$.

An idealized risk argument motivates retaining multiple diagnostic observations. Let $\mathcal O_m=\{o_1,\ldots,o_m\}$ be the observations collected after $m$ interactions and define
\begin{equation}
R_m^\star=\inf_{f\in\mathcal F}
\mathbb E[\ell(\mathbf y,f(\mathbf x,\mathbf u^{[t]},\mathcal O_m))].
\end{equation}
If the predictor family can ignore additional observations, then $R_{m+1}^\star\leq R_m^\star$ for leakage-free tool execution. This is a property of the idealized predictor family, not a guarantee that a finite learned model improves after every interaction. Deterministic diagnostics also need not add external information beyond the origin-available inputs $(\mathbf x,\mathbf u^{[t]})$; they provide alternative representations that can support \mbox{the forecasting model}.

\input{reproducibility}

\section*{Statements and Declarations}

\paragraph{Funding}
This research was supported by grants from the National Natural Science Foundation of China (No.
62502486, 62337001), the grants of Provincial Natural Science Foundation of Anhui Province (No. 2408085QF193), the Fundamental Research Funds for the Central Universities of China (No. WK2150110032).

\paragraph{Competing interests}
The authors have no relevant financial or non-financial interests to disclose.

\paragraph{Data availability}
The sources and access conditions for the benchmark datasets used in this study are documented in Appendix~\ref{app:data_provenance}. CastFlow-specific processed windows and exact split manifests are not publicly archived. Requests for study-specific prepared data should be directed to the corresponding author and will be considered subject to the original providers' access and redistribution conditions.

\paragraph{Code availability}
A reference implementation of CastFlow is publicly available at \url{https://github.com/Forever-Pan/CastFlow}. The archived public revision and its reproducibility scope are documented in Appendix~\ref{app:reproducibility}.

\bibliographystyle{spmpsci_citeorder}
\begingroup
\clubpenalty=3000
\widowpenalty=3000
\brokenpenalty=1000
\let\url\biburl
\bibliography{CastFlow}
\endgroup

\end{document}

%% file: Datasets.tex
\begin{table*}[t!]
    \centering
    \caption{Summary of the benchmark datasets for time series forecasting. "Dim" denotes the total number of recorded columns, including the timestamp column, while the feature list reports only the target variable and covariates. Shared feature lists apply to both ETT rows and both renewable-power rows. All datasets use a 7:1:2 training/validation/test split.}
    \renewcommand{\arraystretch}{1.35}
    \setlength{\tabcolsep}{6pt}
    \label{tab:datasets_summary}
    \newlength{\datasetfeaturewidth}
    \setlength{\datasetfeaturewidth}{\dimexpr\textwidth-14\tabcolsep-7.05cm\relax}
    \begin{tabular}{@{}>{\raggedright\arraybackslash}p{1.3cm}>{\raggedright\arraybackslash}p{1.85cm}>{\centering\arraybackslash}p{1.4cm}>{\centering\arraybackslash}p{.65cm}>{\centering\arraybackslash}p{.5cm}>{\centering\arraybackslash}p{.5cm}>{\centering\arraybackslash}p{.85cm}>{\raggedright\arraybackslash}p{\datasetfeaturewidth}@{}}
    \toprule
    \multicolumn{1}{c}{Dataset} & Domain & Frequency & Dim & $L$ & $H$ & Stride & \multicolumn{1}{c}{Features (Target \& Covariates)} \\
    \midrule
    BE & Electricity Market & 1h & 4 & 168 & 24 & 48 & Electricity price, Generation forecast, System load forecast \\
    DE & Electricity Market & 1h & 4 & 168 & 24 & 48 & Electricity price, Wind power forecast, Amprion zonal load forecast \\
    NP & Electricity Market & 1h & 4 & 168 & 24 & 48 & Electricity price, Grid load forecast, Wind power forecast \\
    FR & Electricity Market & 1h & 4 & 168 & 24 & 48 & Electricity price, Generation forecast, System load forecast \\
    PJM & Electricity Market & 1h & 4 & 168 & 24 & 48 & Electricity price, System load forecast, Zonal COMED load forecast \\
    \midrule
    ETTh & Power Grid & 1h & 8 & 96 & 96 & 48 & Oil temperature, High useful load, High useless load, Middle useful load, Middle useless load, Low useful load, Low useless load \\
    ETTm & Power Grid & 15min & 8 & 96 & 96 & 96 & \\
    \midrule
    WP & Renewable Energy Generation & 15min & 8 & 96 & 96 & 96 & Real power, Direct radiation, Wind direction 80m, Wind speed 80m, Temperature 2m, Relative humidity 2m, Precipitation \\
    SP & Renewable Energy Generation & 15min & 8 & 96 & 96 & 96 & \\
    \midrule
    MOPEX & Streamflow & 1day & 6 & 96 & 96 & 48 & Streamflow discharge, Mean areal precipitation, Climatic potential evaporation, Maximum air temperature, Minimum air temperature \\
    \bottomrule
    \end{tabular}
\end{table*}

%% file: main_table.tex
\begin{table*}[t!]
    \centering
    \caption{MSE and MAE on ten forecasting benchmarks under cross-domain joint training. The two panels report short- and long-term tasks for the same models. Bold and underlined values indicate the best and second-best results, respectively.}
    \label{tab:main_results}
    \setlength{\tabcolsep}{6pt}
    \renewcommand{\arraystretch}{1}
    \begin{tabular*}{\textwidth}{@{\extracolsep{\fill}}l*{10}{r}@{}}
    \toprule
    \multicolumn{11}{c}{Short-term forecasting} \\
    \cmidrule(lr){1-11}
    Model & \multicolumn{2}{c}{BE} & \multicolumn{2}{c}{DE} & \multicolumn{2}{c}{NP} & \multicolumn{2}{c}{FR} & \multicolumn{2}{c}{PJM} \\
    & MSE & MAE & MSE & MAE & MSE & MAE & MSE & MAE & MSE & MAE \\
    \midrule
    Prophet & 992.90 & 16.94 & 320.63 & 13.25 & 57.79 & 5.00 & 1035.70 & 14.42 & 52.26 & 5.53 \\
    ARIMA & 869.14 & 17.04 & 372.91 & 11.82 & 55.51 & 4.45 & 964.00 & 13.22 & 33.02 & 4.00 \\
    XGBoost & 1344.39 & 16.44 & 227.01 & 10.21 & 28.31 & 3.24 & 1044.82 & 10.14 & 30.36 & 3.86 \\
    LightGBM & 1168.62 & 15.71 & 231.44 & \underline{9.57} & 30.92 & 3.38 & 933.84 & 9.96 & 29.63 & 3.88 \\
    Autoformer & 891.02 & 16.62 & 331.37 & 13.33 & 46.11 & 4.86 & 934.55 & 13.37 & 77.59 & 6.43 \\
    DLinear & 658.66 & 12.84 & 239.98 & 12.84 & 32.22 & 3.84 & 811.62 & 10.45 & 42.16 & 4.65 \\
    PatchTST & 627.27 & 11.44 & \underline{208.93} & 10.04 & 24.64 & 3.27 & \underline{797.42} & 9.13 & 31.88 & 4.17 \\
    iTransformer & \underline{606.71} & 11.25 & 230.05 & 10.23 & 27.10 & 3.44 & 940.51 & 11.10 & 35.14 & 4.37 \\
    TimeXer & 703.00 & 10.70 & 252.05 & 10.26 & 27.31 & 3.44 & 834.86 & 9.38 & \underline{25.88} & 3.65 \\
    ConvTimeNet & 621.73 & 10.98 & 231.26 & 10.80 & 28.13 & 3.62 & 882.85 & 8.97 & 28.71 & 3.79 \\
    Chronos & 625.88 & \underline{9.63} & 223.24 & 10.00 & 24.39 & 3.21 & 812.40 & \underline{8.24} & \textbf{25.70} & \underline{3.63} \\
    TimesFM & 784.98 & 10.92 & 219.57 & 9.98 & 25.11 & 3.20 & 870.17 & 8.84 & 33.99 & 4.14 \\
    Sundial & 651.50 & 11.00 & 264.72 & 10.71 & 28.68 & 3.36 & 942.46 & 8.54 & 30.72 & 3.99 \\
    Time-LLM & 729.68 & 11.88 & 253.97 & 10.85 & 33.05 & 3.71 & 936.46 & 11.51 & 35.84 & 4.32 \\
    PromptCast & 824.27 & 13.38 & 293.24 & 10.93 & 43.16 & 3.93 & 1105.16 & 14.42 & 47.62 & 5.58 \\
    TokenCast & 654.38 & 11.48 & 232.21 & 10.84 & 31.49 & 3.59 & 963.09 & 11.82 & 32.37 & 3.92 \\
    S$^2$IP-LLM & 701.24 & 11.89 & 313.82 & 11.15 & 29.67 & 3.42 & 872.55 & 10.93 & 35.67 & 4.16 \\
    TimeReasoner & 691.28 & 11.62 & 272.19 & 10.77 & 31.27 & 3.62 & 899.47 & 11.01 & 34.38 & 4.22 \\
    Time-R1 & 635.23 & 10.23 & 210.28 & 9.89 & 26.28 & 3.35 & 842.19 & 10.92 & 27.35 & 3.80 \\
    TimeSeriesScientist & 906.68 & 14.94 & 245.40 & 10.39 & 46.31 & 4.16 & 904.65 & 12.71 & 45.40 & 5.39 \\
    AlphaCast & 653.98 & 10.71 & 248.98 & 10.23 & \underline{24.16} & \underline{3.19} & 804.89 & 9.10 & 26.53 & \textbf{3.59} \\
    \midrule
    CastFlow & \textbf{546.87} & \textbf{9.60} & \textbf{200.47} & \textbf{9.50} & \textbf{23.92} & \textbf{3.11} & \textbf{707.96} & \textbf{8.16} & 27.45 & 3.74 \\
    \bottomrule
    \end{tabular*}
    \par\medskip
    \begin{tabular*}{\textwidth}{@{\extracolsep{\fill}}l*{10}{r}@{}}
    \toprule
    \multicolumn{11}{c}{Long-term forecasting} \\
    \cmidrule(lr){1-11}
    Model & \multicolumn{2}{c}{ETTh} & \multicolumn{2}{c}{ETTm} & \multicolumn{2}{c}{WP} & \multicolumn{2}{c}{SP} & \multicolumn{2}{c}{MOPEX} \\
    & MSE & MAE & MSE & MAE & MSE & MAE & MSE & MAE & MSE & MAE \\
    \midrule
    Prophet & 46.70 & 4.59 & 21.18 & 3.11 & 8071.24 & 56.06 & 64.81 & 6.44 & 12.40 & 2.50 \\
    ARIMA & 10.88 & 2.43 & 2.71 & 1.17 & 2106.64 & 28.77 & 80.48 & 4.60 & 9.19 & 1.85 \\
    XGBoost & 13.37 & 2.96 & 3.41 & 1.37 & 2384.63 & 37.99 & 33.85 & 3.50 & 6.91 & 1.68 \\
    LightGBM & 12.22 & 2.75 & 2.91 & 1.26 & 2290.27 & 34.81 & 29.50 & 2.80 & 9.63 & 1.84 \\
    Autoformer & 11.60 & 2.64 & 4.23 & 1.57 & 2568.13 & 37.01 & 39.14 & 4.34 & 6.17 & 1.96 \\
    DLinear & 8.51 & 2.24 & 2.63 & 1.14 & \underline{1932.59} & 29.33 & 19.06 & 2.74 & 5.26 & 1.33 \\
    PatchTST & 8.40 & 2.17 & 2.62 & 1.11 & 2270.10 & 32.47 & 19.44 & 2.78 & 5.36 & 1.44 \\
    iTransformer & 8.31 & 2.14 & 3.14 & 1.28 & 2064.60 & 29.90 & 17.96 & \underline{2.67} & 4.88 & 1.34 \\
    TimeXer & 8.77 & 2.20 & 2.56 & 1.11 & 2170.25 & 31.92 & 20.12 & 2.92 & \underline{4.83} & \underline{1.28} \\
    ConvTimeNet & 8.62 & 2.18 & 2.71 & 1.16 & 2081.37 & \underline{28.52} & 23.22 & 3.10 & 5.07 & 1.29 \\
    Chronos & 9.40 & 2.25 & 2.75 & 1.18 & 2269.43 & 29.63 & 79.28 & 4.57 & 5.29 & 1.31 \\
    TimesFM & 9.20 & 2.26 & 2.78 & 1.15 & 2228.19 & 29.84 & 45.91 & 3.37 & 5.58 & 1.35 \\
    Sundial & 9.44 & 2.32 & 3.21 & 1.24 & 2168.81 & 31.22 & 100.40 & 6.61 & 5.15 & 1.42 \\
    Time-LLM & 8.04 & 2.08 & 2.57 & 1.13 & 2833.24 & 38.05 & \underline{17.25} & 2.68 & 5.77 & 1.44 \\
    PromptCast & 9.46 & 2.82 & 4.25 & 1.60 & 2626.89 & 36.20 & 18.21 & 2.84 & 6.70 & 1.59 \\
    TokenCast & 8.28 & 2.17 & 2.50 & \underline{1.10} & 2521.39 & 34.29 & 18.13 & 2.74 & 5.52 & 1.39 \\
    S$^2$IP-LLM & 9.68 & 2.39 & 3.01 & 1.29 & 2497.38 & 33.13 & 18.46 & 2.96 & 6.29 & 1.58 \\
    TimeReasoner & 9.14 & 2.18 & 4.04 & 1.57 & 2228.24 & 31.57 & 20.91 & 2.98 & 5.92 & 1.46 \\
    Time-R1 & \underline{8.02} & \underline{2.04} & 2.80 & 1.11 & 2054.88 & 30.13 & 17.42 & \underline{2.67} & 5.83 & 1.35 \\
    TimeSeriesScientist & 9.09 & 2.24 & 5.28 & 1.61 & 2551.19 & 33.59 & 25.82 & 3.14 & 5.60 & 1.42 \\
    AlphaCast & 8.47 & 2.21 & \underline{2.48} & 1.16 & 2131.02 & 30.22 & 18.89 & 2.80 & 5.19 & 1.43 \\
    \midrule
    CastFlow & \textbf{8.00} & \textbf{2.03} & \textbf{2.36} & \textbf{1.04} & \textbf{1719.99} & \textbf{27.40} & \textbf{16.90} & \textbf{2.50} & \textbf{3.60} & \textbf{1.19} \\
    \bottomrule
    \end{tabular*}
\end{table*}

%% file: Ablation-1.tex
\begin{table*}[t]
    \centering
    \caption{Component ablation under the evaluated configurations. ``w/o Toolkit'' removes the ensemble prior and diagnostics together with dependent tool planning; ``w/o Memory'' removes experience retrieval; ``w/o Refl.'' removes self-correction. The complete configuration has the lowest reported errors in this table.}
    \label{tab:ablation_full}
    
    \setlength{\tabcolsep}{6pt} 
    \renewcommand{\arraystretch}{1.11}
    \begin{tabular*}{\textwidth}{@{\extracolsep{\fill}}c|l|ccccc|ccccc@{}}
        \toprule
        \multicolumn{2}{c|}{Settings} & \multicolumn{5}{c|}{Short-term Forecasting} & \multicolumn{5}{c}{Long-term Forecasting} \\
        \cmidrule(r){1-2} \cmidrule(lr){3-7} \cmidrule(l){8-12}
        Metrics & \multicolumn{1}{c|}{Variants} & BE & DE & NP & FR & PJM & ETTh & ETTm & WP & SP & MOPEX \\
        \midrule
        \multirow{4}{*}{MSE} & w/o Refl. & 1027.76 & 394.83 & 38.64 & 801.89 & 45.06 & 33.84 & 20.51 & 5102.29 & 22.72 & 5.18 \\
         & w/o Toolkit & 836.24 & 358.33 & 31.21 & 724.87 & 42.33 & 23.71 & 17.68 & 3517.03 & 21.46 & 4.54 \\
         & w/o Memory & 551.54 & 204.54 & 24.67 & 712.30 & 28.30 & 8.45 & 2.41 & 1820.92 & 17.49 & 4.63 \\
         & Full Model & \textbf{546.87} & \textbf{200.47} & \textbf{23.92} & \textbf{707.96} & \textbf{27.45} & \textbf{8.00} & \textbf{2.36} & \textbf{1719.99} & \textbf{16.90} & \textbf{3.60} \\
        \midrule
        \multirow{4}{*}{MAE} & w/o Refl. & 15.35 & 12.44 & 4.07 & 10.55 & 5.47 & 5.37 & 4.27 & 43.48 & 3.85 & 1.79 \\
         & w/o Toolkit & 13.53 & 11.50 & 3.62 & 9.46 & 4.77 & 3.55 & 2.86 & 37.52 & 2.96 & 1.36 \\
         & w/o Memory & 9.62 & 9.77 & 3.14 & 8.22 & 3.79 & 2.18 & 1.08 & 28.69 & 2.55 & 1.38 \\
         & Full Model & \textbf{9.60} & \textbf{9.50} & \textbf{3.11} & \textbf{8.16} & \textbf{3.74} & \textbf{2.03} & \textbf{1.04} & \textbf{27.40} & \textbf{2.50} & \textbf{1.19} \\
        \bottomrule
    \end{tabular*}
\end{table*}

%% file: toolkit_ablation.tex
\begin{table*}[t]
    \centering
    \caption{Tool-category ablation. The anchorer supplies the ensemble prior, while the profiler and monitor supply diagnostics. The residual diagnoser (Diagn.) is available only during training. Best reported values are bolded.}
    \label{tab:toolkit_ablation}
    \setlength{\tabcolsep}{6pt}
    \renewcommand{\arraystretch}{1.11}
        \begin{tabular*}{\textwidth}{@{\extracolsep{\fill}}c|l|ccccc|ccccc@{}}
            \toprule
            \multicolumn{2}{c|}{Settings} & \multicolumn{5}{c|}{Short-term Forecasting} & \multicolumn{5}{c}{Long-term Forecasting} \\
            \cmidrule(r){1-2} \cmidrule(lr){3-7} \cmidrule(l){8-12}
            Metrics & \multicolumn{1}{c|}{Variants} & BE & DE & NP & FR & PJM & ETTh & ETTm & WP & SP & MOPEX \\
            \midrule
            \multirow{5}{*}{MSE}
            & w/o Anchorer & 697.35 & 270.36 & 28.50 & 786.70 & 47.89 & 17.30 & 3.53 & 2718.42 & 20.17 & 5.67 \\
            & w/o Profiler & 567.03 & 201.87 & 24.39 & \textbf{675.67} & 28.15 & \textbf{7.88} & \textbf{2.33} & 1821.36 & 17.42 & 4.69 \\
            & w/o Monitor & 547.23 & 200.90 & 24.66 & 695.90 & 28.57 & 7.90 & 2.43 & 1890.65 & 19.38 & 4.52 \\
            & w/o Diagn. & 553.86 & 201.67 & 24.48 & 710.00 & 28.18 & 8.08 & 2.38 & 1805.12 & 17.49 & 4.54 \\
            & Full Model & \textbf{546.87} & \textbf{200.47} & \textbf{23.92} & 707.96 & \textbf{27.45} & 8.00 & 2.36 & \textbf{1719.99} & \textbf{16.90} & \textbf{3.60} \\
            \midrule
            \multirow{5}{*}{MAE}
            & w/o Anchorer & 13.37 & 11.16 & 3.53 & 10.55 & 4.85 & 3.00 & 1.36 & 35.05 & 2.62 & 1.35 \\
            & w/o Profiler & 9.71 & 9.68 & 3.13 & \textbf{7.59} & 3.78 & 2.06 & 1.05 & 28.40 & 2.51 & 1.37 \\
            & w/o Monitor & 9.62 & 9.66 & 3.15 & 7.88 & 3.82 & 2.04 & 1.08 & 28.96 & \textbf{2.45} & 1.36 \\
            & w/o Diagn. & 9.68 & 9.69 & 3.12 & 8.20 & 3.78 & 2.07 & 1.06 & 28.39 & 2.53 & 1.36 \\
            & Full Model & \textbf{9.60} & \textbf{9.50} & \textbf{3.11} & 8.16 & \textbf{3.74} & \textbf{2.03} & \textbf{1.04} & \textbf{27.40} & 2.50 & \textbf{1.19} \\
            \bottomrule
        \end{tabular*}
\end{table*}

%% file: reward.tex
\begin{table}[t]
    \centering
    \caption{Reward-design comparison for the specialized forecasting model. Hybrid MSE combines absolute forecasting accuracy with improvement over the ensemble prior. Abs. and Rel. denote absolute and relative rewards. Best reported values are bolded.}
    \label{tab:reward_ablation_final}
    \setlength{\tabcolsep}{6pt}
    \renewcommand{\arraystretch}{1.20}
    \begin{tabular*}{\columnwidth}{@{\extracolsep{\fill}}llrrrr@{}}
    \toprule
    Data & Metric & \shortstack{Abs.\\MSE} & \shortstack{Rel.\\MSE} & \shortstack{Hybrid\\MSE} & \shortstack{Abs.\\MAE} \\
    \midrule
    \multirow{2}{*}{BE} & MSE & 549.38 & 551.78 & \textbf{546.87} & 552.55 \\
     & MAE & 9.68 & 9.61 & \textbf{9.60} & 9.62 \\
    \midrule
    \multirow{2}{*}{DE} & MSE & 201.54 & 202.85 & \textbf{200.47} & 205.13 \\
     & MAE & \textbf{9.49} & 9.54 & 9.50 & 9.80 \\
    \midrule
    \multirow{2}{*}{SP} & MSE & 17.26 & 17.07 & \textbf{16.90} & 17.19 \\
     & MAE & 2.52 & 2.52 & \textbf{2.50} & 2.52 \\
    \midrule
    \multirow{2}{*}{ETTh} & MSE & 8.09 & 8.10 & \textbf{8.00} & 8.02 \\
     & MAE & 2.08 & 2.08 & \textbf{2.03} & 2.08 \\
    \bottomrule
    \end{tabular*}
\end{table}

%% file: system_recovery.tex
\begin{table*}[t]
    \centering
    \caption{Mean execution time (seconds) for restart and stage recovery on six paired windows per dataset, under normal execution and a controlled post-forecast interruption. All attempts, including failed forecasts, remain in the averages. Timing includes journal I/O and excludes model-service initialization. Bold indicates the lower mean within each condition.}
    \label{tab:system_recovery}
    \fontsize{9}{11}\selectfont
    \setlength{\tabcolsep}{6pt}
    \renewcommand{\arraystretch}{1.11}
    \newlength{\recoverycontentwidth}
    \setlength{\recoverycontentwidth}{\dimexpr\textwidth-14\tabcolsep-2\arrayrulewidth\relax}
    \newlength{\recoverydatasetwidth}
    \setlength{\recoverydatasetwidth}{\dimexpr(\recoverycontentwidth-.22\recoverycontentwidth-.13\recoverycontentwidth)/6\relax}
    \begin{tabular}{@{}>{\centering\arraybackslash}p{.22\recoverycontentwidth}|>{\centering\arraybackslash}p{.13\recoverycontentwidth}|*{6}{>{\centering\arraybackslash}p{\recoverydatasetwidth}}@{}}
        \toprule
        Condition & Executor & BE & DE & ETTh & ETTm & WP & SP \\
        \midrule
        \multirow{2}{*}{Normal} & Restart & 15.02 & 17.20 & \textbf{40.48} & \textbf{41.89} & 39.75 & 41.54 \\
         & Resume & \textbf{14.77} & \textbf{16.72} & 40.74 & 42.67 & \textbf{38.90} & \textbf{40.91} \\
        \midrule
        \multirow{2}{*}{Interrupted} & Restart & 29.01 & 34.30 & 80.41 & 77.30 & 76.22 & 82.56 \\
         & Resume & \textbf{14.80} & \textbf{16.82} & \textbf{40.44} & \textbf{46.28} & \textbf{38.35} & \textbf{46.67} \\
        \bottomrule
    \end{tabular}
\end{table*}

%% file: tool_specification.tex
\begin{table*}[t]
\centering
\caption{Tool inventory, inputs, outputs, and execution scope.}
\label{tab:evidence_tools}
\fontsize{9}{11}\selectfont
\setlength{\tabcolsep}{6pt}
\renewcommand{\arraystretch}{1.06}
\newlength{\toolcontentwidth}
\setlength{\toolcontentwidth}{\dimexpr\textwidth-6\tabcolsep\relax}
\begin{tabular}{@{}>{\raggedright\arraybackslash}p{0.21\toolcontentwidth}>{\raggedright\arraybackslash}p{0.20\toolcontentwidth}>{\raggedright\arraybackslash}p{0.39\toolcontentwidth}>{\raggedright\arraybackslash}p{0.20\toolcontentwidth}@{}}
\toprule
Tool & Input & Main output & Invocation / phase \\
\midrule
\multicolumn{4}{@{}l}{\textbf{Foundational anchorer (1 tool)}} \\
Model auxiliary & Lookback; historical expert library & Ensemble forecast prior $\hat{\mathbf y}_{\mathrm{base}}$ & Mandatory; training and inference \\
\addlinespace[3pt]
\multicolumn{4}{@{}l}{\textbf{Statistical and spectral profiler (4 tools)}} \\
Statistical analysis & Target lookback & Mean, standard deviation, extrema, and range & Optional; training and inference \\
Basic statistics & Target lookback & Median, MAD, and spectral entropy & Optional; training and inference \\
Data quality & Lookback values and missingness & Dropout ratio; bounds $[\mu-\kappa\sigma,\mu+\kappa\sigma]$ & Optional; training and inference \\
Comprehensive feature & Target lookback & Aggregated statistical, spectral, and quality diagnostics & Optional; training and inference \\
\addlinespace[3pt]
\multicolumn{4}{@{}l}{\textbf{Dynamics monitor (5 tools)}} \\
Trend analysis & Target lookback & Linear slope and trend direction & Optional; training and inference \\
Changepoint trend & Target lookback & First- and second-difference summaries & Optional; training and inference \\
Cross-channel & Target and auxiliary lookback & Historical correlations and lead--lag associations & Optional; training and inference \\
Exogenous analysis & Target and auxiliary lookback & Historical target--auxiliary correlations & Mandatory; training and inference \\
Event summary & Target lookback & Rise, fall, flat, or oscillation label & Optional; training and inference \\
\addlinespace[3pt]
\multicolumn{4}{@{}l}{\textbf{Residual diagnoser (1 tool)}} \\
Autoregressive residual & Target lookback & Lookback-fitted residual mean and lag-one autocorrelation & Optional; training only \\
\bottomrule
\end{tabular}
\par\smallskip
\begin{minipage}{\textwidth}
\footnotesize
Inputs must be available at the forecast origin. Horizon covariate forecasts $\mathbf u^{[t]}$ enter the forecasting context through the supported interface; historical associations use lookback observations. Optional tools are selected by the frozen planner. The residual tool is skipped at inference; archived residuals describe training windows. Diagnostic features may overlap across categories. MAD denotes median absolute deviation.
\end{minipage}
\end{table*}

%% file: reproducibility.tex
\section{Reproducibility and Information Availability}
\label{app:reproducibility}

\subsection{Forecast-Origin Information Boundary}
Equation~\ref{eq:available_information} defines admissible source information $\mathcal I_t$ at forecast origin $t$. The inference protocol supplies the observed lookback and horizon-aligned covariate forecasts $\mathbf u^{[t]}$ issued by $t$; diagnostics use admissible inputs and fixed offline resources. An issued forecast can describe a future period, whereas its realized observation is unavailable at $t$. Supplied forecasts remain fixed through reflection retries, excluding later revisions. Admissibility therefore depends on provenance and release time, not merely the time described \mbox{by a value}.

Test Experience Memory contains training experiences and remains read-only; training targets support offline trajectory selection. The current test target is used only after forecasting for evaluation, never for retrieval, planning, reflection, or memory maintenance. Request history $\mathcal H_j$ may change through retries. Archived diagnostics remain examples from their training windows, not observations of \mbox{the test window}.

\subsection{Reported Experimental Configuration}
Table~\ref{tab:reported_configuration} summarizes the reported settings. The anchorer expert pool, comparison baselines, and forecasting LLM are separate configurations; historical expert-weighting losses and response-scoring MSE also \mbox{serve different roles}.

\begin{table}[t]
\centering
\caption{Reported settings for the main forecasting evaluation. Memory is fixed throughout \mbox{test inference}.}
\label{tab:reported_configuration}
\fontsize{9}{11}\selectfont
\setlength{\tabcolsep}{3pt}
\renewcommand{\arraystretch}{1.12}
\begin{tabular}{@{}>{\raggedright\arraybackslash}p{0.35\columnwidth}>{\raggedright\arraybackslash}p{0.60\columnwidth}@{}}
\toprule
Component & Setting \\
\midrule
General-purpose model & Grok 4; planning and reflection frozen \\
Specialized model & Qwen3-4B; forecasting parameters updated \\
SFT & Learning rate $5\times10^{-5}$; batch size 8; 1 epoch \\
RLVR & GRPO; learning rate $2\times10^{-6}$; 3 epochs \\
GRPO sampling & Group size 8; temperature 1.0; KL coefficient 0 \\
GRPO scoring & Direct response-level reward; no candidate-wise reflection loop \\
Completion limit & 5,000 tokens for the trainable forecasting model \\
Short-term tasks & Lookback 168; forecast horizon 24 \\
Long-term tasks & Lookback 96; forecast horizon 96 \\
Data partition & Chronological training/validation/test ratio 7:1:2 \\
Horizon covariates & Auxiliary forecasts issued no later than the forecast origin \\
Test Experience Memory & Training-built library; retrieval only \\
Evaluation scale & MSE and MAE on the original target scale \\
\bottomrule
\end{tabular}
\end{table}

\subsection{Configuration Scope}
\label{sec:result_configuration}
The GRPO sampling settings describe training, whereas greedy decoding and dataset-specific checkpoints apply to the recovery experiment in Section~\ref{sec:system_evaluation}.

\subsection{Prompt and Response Interfaces}
Planning receives the observed window, tool descriptions, mandatory-tool requirements, and retrieved guidance, and returns a tool schedule. Forecasting receives observations, $\mathbf u^{[t]}$, the ensemble prior, diagnostics, procedural guidance, and revision feedback, and returns reasoning plus horizon-wide predictions. Memory therefore guides both planning and forecasting; archived observations and answers remain training examples. Relevant tools consume covariate forecasts through their supported interfaces. Reflection first checks structural validity, then evidence consistency without test targets; semantic acceptance does not guarantee \mbox{numerical accuracy}.

Public prompt examples are in \path{forecasting/modules.py} and \path{forecasting/prompts.py} at revision \href{https://github.com/Forever-Pan/CastFlow/tree/c6a993d681f934da0f5cdff644b08388ca7b3726}{\texttt{c6a993d681f9}}.

\subsection{Training Samples and Inference Outputs}
\label{app:training_output_protocol}
SFT imitates teacher responses in selected pairs $(\mathcal C,r^*)$; memory selection instead retains historical procedures. RLVR samples responses sharing a context, diagnostics, covariate forecasts, and prior, and scores each numerical answer directly. Groupwise advantages train only specialized forecasting-response tokens, without candidate-wise reflection or retries or gradients to planning \mbox{and reflection tokens}.

If the retry budget is exhausted without acceptance, fallback uses a structurally valid request-history response, then the ensemble prior; an output is missing or abnormal only if neither is usable. Fallback never selects by test error.

\subsection{Reward Configuration}
\label{app:reward_configuration}
Equations~\ref{eq:absolute_forecast_reward} and~\ref{eq:forecast_reward} define the absolute and relative rewards, with $\lambda=5$, $\delta=0.5$, and Table~\ref{tab:reward_scales}'s fixed MSE scales. Here $\gamma_d$ controls the absolute mapping and $\nu_d$ scales prior-minus-candidate MSE. Without a usable matched prior, only the absolute \mbox{term is returned}.

\begin{table}[t]
\centering
\caption{Dataset-specific reward scales (main configuration).}
\label{tab:reward_scales}
\fontsize{9}{11}\selectfont
\setlength{\tabcolsep}{8pt}
\begin{tabular}{@{}lrr@{}}
\toprule
Dataset & Absolute scale $\gamma_d$ & Relative scale $\nu_d$ \\
\midrule
BE & 800 & 10 \\
DE & 350 & 10 \\
FR & 1000 & 10 \\
NP & 40 & 3 \\
PJM & 45 & 4 \\
ETTh & 12 & 2 \\
ETTm & 3.5 & 1 \\
WP & 3000 & 30 \\
SP & 50 & 3 \\
MOPEX & 7 & 1.5 \\
\bottomrule
\end{tabular}
\end{table}

Before numerical scoring, an unparseable nonempty answer receives $-1$; an empty or shorter-than-$H$ answer receives $-0.8$. For $n>H$ values, the reward is
\begin{equation}
\max\!\left(-0.5-0.6\frac{n-H}{H},-0.8\right).
\end{equation}
A missing or unparseable target or NaN candidate MSE receives $-1$; NaN baseline MSE disables only the relative term. Numerical training reward does not require semantic reflection acceptance. The reference parser extracts prediction and target sequences from supported keys, and the horizon-wide losses in Section~\ref{sec:optimization} assume \mbox{target-aligned predictions}.

\subsection{Dataset Provenance and Access}
\label{app:data_provenance}
The five EPF markets originate from the benchmark of Lago et al.~\cite{lago2021epf}; this study uses the \href{https://github.com/thuml/TimeXer/tree/main/dataset/EPF}{TimeXer EPF collection}~\cite{wang2024timexer}. ETT data are provided through the authors' \href{https://github.com/zhouhaoyi/ETDataset/tree/main/ETT-small}{ETDataset repository}. The renewable-power datasets originate from the iFLYTEK challenge cited in the main text; access remains subject to the \href{https://challenge.xfyun.cn/topic/info?type=renewable-power-forecast\&option=ssgy\&ch=dwsf259}{competition platform}. MOPEX provenance is described by Schaake et al.~\cite{schaake2006us}; the \mbox{\href{https://iahs.info/uploads/dms/13600.04-9-28-SCHAAKE.pdf}{original data description}} and its \mbox{\href{https://digital.library.unt.edu/ark:/67531/metadc883012/}{institutional report archive}} document the source and \mbox{collection procedure}.

The experimental settings specify chronological splits, targets, horizons, and metric scale. Study-specific processed windows and split manifests are not publicly archived, as stated in Data availability.
\FloatBarrier